%% file: Multi-DoF.tex
\RequirePackage{amsmath} 

\documentclass{article}
\providecommand{\keywords}[1]{\textbf{Keywords:} #1}
\usepackage[english]{babel}
\usepackage{bm,upgreek}
\usepackage{amssymb}
\usepackage[a4paper, left=2cm, right=2cm, top=2cm]{geometry}
\usepackage{graphicx}
\usepackage{booktabs}
\usepackage[margin=10pt,font=small,labelfont=bf,labelsep=endash]{caption}
\usepackage{authblk}
\usepackage[breaklinks=true,
            hidelinks,
            pdfauthor={Stephanie Kirmse},
            pdftitle={Multi-DoF},
            pdfkeywords={??, ??, ??}]{hyperref}

\usepackage[backend=biber,citestyle=numeric-comp,bibstyle=ieee,dashed=false,url=false]{biblatex}
\AtEveryBibitem{\ifentrytype{article}{\clearfield{note}}{}}
\AtEveryBibitem{\ifentrytype{book}{\clearfield{note}}{}}
\AtEveryBibitem{\ifentrytype{article}{\clearfield{issn}}{}}
\DeclareCiteCommand{\cite}
    {\usebibmacro{cite:init}%
     \usebibmacro{prenote}}
    {\usebibmacro{citeindex}%
         \ifnumequal{\value{citecount}}{1}{\printtext[bibhyperref]{[}}{}%
        \usebibmacro{cite:comp}%
    }
    {}
    {\usebibmacro{cite:dump}%
     \ifnumequal{\value{citecount}}{\value{citetotal}}{\printtext[bibhyperref]{]}}{}%
     \usebibmacro{postnote}%
    }
\usepackage{amsmath}
\usepackage{tabularx} 
\usepackage{color}
\usepackage{hyperref}

\begin{document}
    \input{sec_0_1_Front_Page.tex}
    \input{sec_0_2_Abstract.tex}

    \input{sec_1_0_Introduction.tex}
    \input{sec_2_0_Overview.tex}
    \input{sec_3_0_Design_Problem.tex}
    \input{sec_4_0_Optimization_Formulation.tex}
    \input{sec_5_0_Design_Examples.tex}

    \input{sec_6_0_Conclusion_and_future_work.tex}

    \printbibliography
\end{document}

%% file: sec_0_1_Front_Page.tex
\title{Topology-optimization based design of multi-degree-of-freedom compliant mechanisms (mechanisms with multiple pseudo-mobility)}

\author[1]{Stephanie Kirmse}
\author[1]{Lucio Flavio Campanile}
\author[1]{Alexander Hasse (corresponding author)}

\affil[1]{Chemnitz University of Technology, Professorship Machine Elements and Product Development, Chemnitz, Germany}

\date{}

\maketitle

%% file: sec_0_2_Abstract.tex
\begin{abstract}
Unlike conventional mechanisms, compliant mechanisms produce the desired deformations by exploiting elastic strain and do not need, therefore, moving parts. The number of degrees of freedom of a conventional mechanism, also called mobility, is the number of independent coordinates needed to define a configuration of the mechanism. Due to the different operating principle, such definition of degree of freedom or mobility cannot be directly applied to compliant mechanisms. While those terms are not able to denote a property of a given compliant mechanism, they are meaningful when applied to the design of a compliant mechanism. Compliant mechanisms are, however, mostly seen as elastic structures, for which the term degree of freedom is used in a different meaning. In order to avoid ambiguities, the term pseudo-mobility (already introduced in previous published work) will be used to denote the number of scalar parameters needed to identify one single desired deformation, i.e. one single deformation for which the compliant mechanism is designed. Many synthesis approaches exist for compliant mechanisms with single pseudo-mobility (commonly referred to as "single degree of freedom mechanisms"). In the case of compliant mechanisms with multiple pseudo-mobility (multiple-degree of freedom mechanisms), only synthesis approaches for relatively simple mechanisms exist so far, while systems for more complex tasks like shape adaptation are not covered. In addition, only certain cases of transverse loads are included in the synthesis with these approaches. In this paper, a novel optimization algorithm is presented that addresses these two shortcomings. The algorithm is tested on a simple mechanism with one translation and one rotation kinematic degree of freedom, a compliant parallel mechanism for pure translation and a shape-adaptive structure.
\end{abstract}

\keywords{compliant mechanisms, topology optimization, selective compliance, multi-degree-of-freedom, compliant parallel mechanism, design of compliant mechanisms}

%% file: sec_1_0_Introduction.tex
\section{Introduction}
\label{sec:Introduction}
This paper focuses on the design of so-called multiple-degree of freedom compliant mechanism. We included this terminology in the paper’s title for the sake of uniformity with many published studies of other authors, but we consider it an unfortunate choice. For this reason we decided to begin the paper with a discussion on it, which, by the way, will provide a good introduction to the paper’s content.

There are two main reasons why the mentioned use of the term “degree of freedom” in conjunction with compliant mechanisms is problematic. As a term, it is used incorrectly or at least ambiguously; as a concept, it suffers from the fact of being arbitrary: no number of degrees of freedom can be assigned to a given compliant mechanism in an objective and rigorous way.

Conventional mechanisms can be idealized as systems of rigid bodies (links) connected by coupling elements (joints). For a given mechanism, the number of degrees of freedom (DoF) or mobility $n$ can be uniquely determined as a function of the number of rigid bodies as well as the number and kind of joints \cite{Gogu2005}. After defining as deformation of the mechanism any motion of the links that does not corresponds to a global rigid-body motion, it can be stated that the mobility expresses the number of kinematic quantities that need to be specified in order to determine a particular deformation of the mechanism. Since any single deformation is associated with the choice of $n$ scalar quantities, there are $\infty^n$ possible deformations.

\begin{figure*}[t]
    \centering
    \includegraphics[width=16cm]{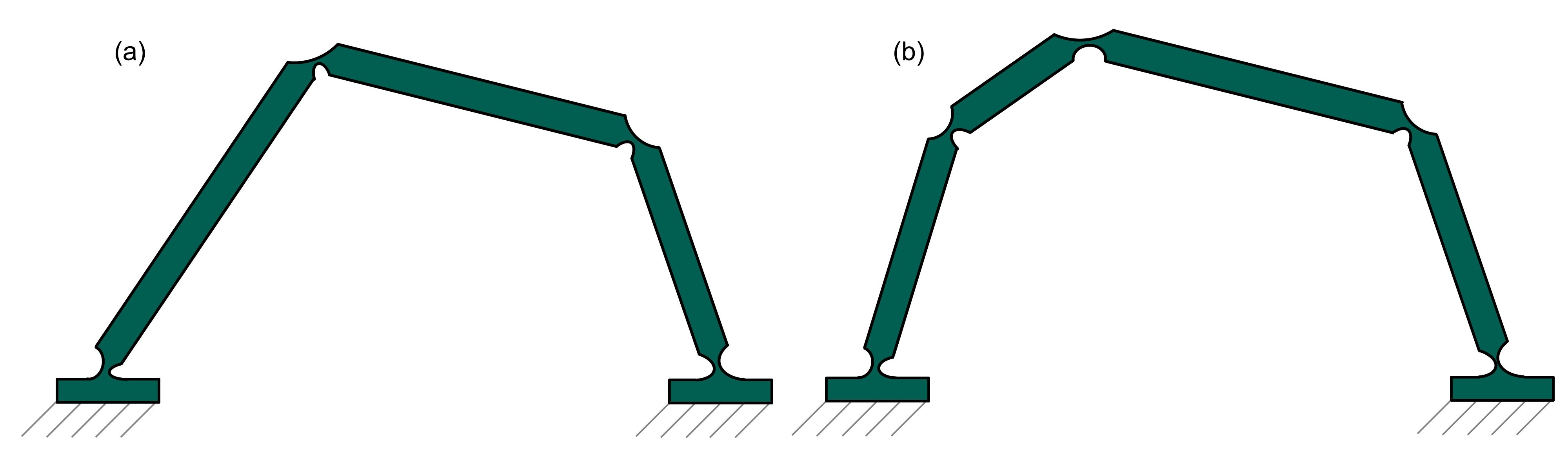}
    \caption{Examples of compliant mechanisms with single pseudo-mobility (a) and double pseudo-mobility (b)}
    \label{figure0}
\end{figure*}

Compliant mechanisms need elastic strain to produce their functional deformations. This imply that they, in the general case, have to be idealized as continua, i.e. as systems with a theoretically infinite number of DoFs. Even when a discretized model is used (e.g. a FE-Model) they usually possess a very large number of DoFs. This also apply for instance to the mechanism of \autoref{figure0} (a), which is, however, commonly seen as a single-DoF compliant mechanism, since it resembles a single-DoF conventional mechanism. Analogously, the mechanism of \autoref{figure0} (b), which is provided with an infinite number of DoFs as well, is denoted as a two-DoF mechanism. This shows the formal part of the problem: the concept of DoFs is used improperly in the “new” (kinematic) sense. Beyond this question of terms, there is a substantial one: The number of “kinematic” DoFs cannot be used as a characteristic of a compliant mechanism, since no way exists to determine such number in an objective and rigorous way for a given mechanism, and so this quantity is, in an analysis framework, arbitrary. 

Two of the authors \cite{Hasse2009} adressed this question for linearized mechanisms on the basis of an eigenvalue analysis. Compliant mechanisms like the ones in \autoref{figure0}, for which a number $m$ of kinematic DoFs (there denoted as pseudo-mobility) is assumed, show a pronounced step between the $m$-th and the $(m+1)$-th eigenvalue of the stiffness matrix. Even on this basis, a quantitative and unambiguous criterion to define the pseudo-mobility of a conventional mechanism requires a convention on how large the increase between two subsequent eigenvalues must be to consider it as “pronounced step”. And such convention would be, in turn, arbitrary.

In a synthesis framework, however, the concept of kinematic DoFs or pseudo-mobility can be used in an objective and exact sense. If a compliant mechanism is designed to mimic a two-DoF conventional mechanism, specific requirements are included in the design procedure to allow $\infty^2$ deformations, each of them assigned to two scalar parameters. In this case, it can be objectively asserted that the mechanism to be designed possesses a pseudo-mobility of two. From now on, we will avoid the term DoFs because of the above described formal ambiguity.

But the pseudo-mobility remains an attribute of the design and not of the mechanism: “the mechanism is designed for a pseudo-mobility of two” is an objective statement, while “the mechanism has a pseudo-mobility of two” is not. In the following, when speaking about a compliant mechanism with a certain value of pseudo-mobility, we will mean that the considered mechanism was designed for that value of pseudo-mobility and/or fulfills corresponding, properly specified design restrictions.

In general, we can say that if a compliant mechanism is designed for $\infty^m$  desired deformations, each of them associated to $m$ scalar parameters, it is designed for a pseudo-mobility of $m$.

In this paper, the \textit{modal procedure} for the topology-optimization based synthesis of compliant mechanisms \cite{Hasse2009, Hasse2016, Kirmse2021} is extended to the case of $m>1$ (multiple pseudo-mobility). Like the original procedure, the presented procedure is valid for the geometrically linear case. The desired deformations can then be expressed as a linear combination of $m$ desired deformation modes.

%% file: sec_2_0_Overview.tex
\section{Overview of the state of the art}
\label{sec:Overview state of the art}

Compliant mechanisms with lumped compliance \cite{Kota1995} of the kind shown in \autoref{figure0} can be efficiently designed on the basis of the pseudo-rigid-body model (PSRM). With the PSRM, the solid-state joints of the compliant mechanism are replaced by classical joints \cite{Howell1994,Howell1996b}. The elastic restoring effect of the deformed material is simulated by a spring that acts between the two links connected through the solid-state joint. With the help of this model, the compliant mechanism can first be designed as a conventional mechanism and then the conventional joints can be converted into suitable solid-state joints \cite{Tang2006,Lu2004,Wang2017}. This approach, called rigid-body replacement method, is one of the two main schools in synthesis of compliant mechanisms. It has the advantage of translating the main part design problem into the design of a conventional mechanism and therefore accessing an established methodology. However, it is strongly limited in the kind of designs it can handle. The position of the center of relative rotation between two members changes, with respect to the members themselves, as a function of the load \cite{Cammarata2016}. In compliant mechanisms with distributed compliance \cite{Kota1995}, where this effect is more pronounced, the replacement with conventional hinges can reveal impracticable.

If the whole potential of compliant mechanisms is to be exploited, with no a-priori restriction in the layout, the second design school, the structural-optimization based approach, is the proper choice. In the structural optimization based approach, synthesis is performed by topology optimization. This distributes material in a previously defined design space according to a given objective function and proper restrictions. In this way, any distribution of flexible and rigid regions compatible with the chosen parameterisation is a possible solution, which makes this method potentially very powerful.
The main disadvantage of the structural-optimization approach is that it does not allow a kinematic design. While the design of conventional mechanisms (and, as a consequence, the rigid-body replacement approach) operates on kinematic constraints which restrict the possible deformations to a clearly defined subset without the need of taking into account the loads acting on the mechanism, the structural-optimization approach operates on the functional relationship between forces and displacements. Every deformation of the mechanism is related to a specific load, which means that the structural-optimization methods are, as a matter of principle, load-case specific.
A subset of the system’s DoF is used to control the mechanism by given forces or displacements (input DoFs). To address all desired deformations, the number of input DoFs must be greater than or equal to $m$. The output DoFs are defined as the ones of relevance for the mechanism’s function and can be subjected to external forces (usually called transverse loads).

A successful optimization formulation (objective function and restrictions) must include a proper number of load cases and can become quite complex. Beyond the pseudo-mobility and the number of input DoFs, the number of output DoFs plays a particular role in this sense, since it determines the number of possible load configurations. Independently on the load-case issue, a large number of output DoF also implies that the desired deformations are defined over a large number of DoF, which must properly be processed by the optimization formulation.
Hence, a large portion of published studies deals with the simplest option, provided by the so-called Single-Input/Single-Output (SISO) mechanisms. Under linear assumptions, just two independent load cases are present (with or without loads at the output port) and the optimization logic can be straightforwardly formulated by maximizing the output displacement under the action of the sole input force and minimizing it (in a multiobjective sense) when the input DoF is blocked and a force acts on the output DoF. In \cite{Frecker1997}, these requirements are translated into an energy criterion that involves the so-called mutual potential energy (MPE), related to the first load case, and the strain energy (SE), related to the second one. Owing to the presence of just one input DoF, a SISO design has a pseudo-mobility of one. 

Another concept which is often used for the synthesis of SISO mechanisms is based on the maximization of the mechanical advantage (MA). This describes the ratio between input and output forces of a compliant mechanism (see e.g. \cite{Sigmund1997}). However, this quantity can only be considered when the mechanism is loaded at the output. The loading results from coupling with a spring of given spring constant. The two load cases described above are implicitly considered in this formulation; the requirement for the lowest possible input force corresponds to maximizing flexibility under no-load conditions, and the requirement for the highest possible output force corresponds to maximizing stiffness under transverse loads. By choosing the spring constant at the output, a weighting of the two goals is obtained. The possibility of combining a transverse load with the no-load case by coupling with a spring is often used in optimization-based synthesis methods. We will use the term "load springs" in this context in the following.

Another function used in SISO structural optimization design approaches is the geometric advantage (GA), which is described as the ratio between the input and the output displacement of the mechanism. It is maximized, for example, in \cite{Lau2001}.

When leaving the SISO-terrain, the availability of structural-optimization based procedures becomes quite sparse. For mechanisms with single pseudo-mobility in the geometrically linear case, in \cite{Frecker1999} the ratio of MPE to SE is maximized for one input DoF, where the output displacements are included into a combined MPE by so-called virtual loads, or a weighted sum of the individual MPEs is used. In \cite{Li2014a}, a multi-objective optimization minimizes the "mean compliance" which, like the SE, is a measure of the structural stiffness of the mechanism, while maximizing a weighted sum of the output displacements. Here, transverse loads are included with the help of one load spring applied per output DoF.

Concerning the case of multiple pseudo-mobility, several synthesis approaches exist specifically for the synthesis of compliant parallel mechanisms consisting of a rigid working platform placed centrally in a defined design space and several, similar compliant structures arranged around it. The multiple pseudo-mobility of the considered compliant parallel mechanisms results from the fact that the working platform retains more than one rigid-body DoF. In some approaches, the compliant structures are designed separately and the result is then combined into a complete compliant parallel mechanism \cite{GuoZhanLum2013,Pham2017}. In \cite{GuoZhanLum2013}, the compliant structures are loaded with unit loads to compute a compliance matrix. This compliance matrix refers to the six rigid-body DoFs in space. All entries of the compliance matrix are then integrated into the objective function in such a way that its minimum search maximizes the entries belonging to the desired directions of motion and minimizes the remaining entries in the multi-objective sense. Suitable assembly of the compliant structures computed in this way subsequently results in a compliant parallel mechanism with pseudo-mobility of three. A parallel mechanism with pseudo-mobility of three is also designed in \cite{Pham2017}. Here, the main-diagonal elements of the stiffness matrix are integrated into the objective function. The ratio between the product of the elements belonging to desired directions of motion and the product of the elements belonging to the undesired directions of motion is minimized. In \cite{Zhang2018}, the geometrical advantages of all the combinations between an input and an output DoF are collected in a Jacobi matrix. However, this only applies to the no-load case. In the objective function, the entries of the Jacobi matrix related to the desired motion are maximized or set to a given value and, in addition, the characteristic input and output stiffnesses are minimized. The characteristic input and output stiffnesses are the elements of the stiffness matrix belonging to the main diagonal and associated with the input and output DoFs. This formulation is used for parallel mechanisms with pseudo-mobility of two and three.

Further synthesis approaches were developed for compliant mechanisms of general kind. In \cite{Wang2010}, a multi-objective optimization is performed for the synthesis of a "grip-and-move manipulator" with a pseudo-mobility of two and two output DoFs, which describes on the one hand the size of the working space and on the other hand the maximum motion of the output DoFs in one desired direction by defined applied forces. No transverse loads are assumed. In \cite{Zhu2018}, the weighted sum of the displacements caused by given applied forces is maximized. The approach is tested on compliant mechanisms with pseudo-mobility of two and three. The approach is suitable for synthesizing mechanisms with a number of output DoF equal to the pseudo-mobility. The transverse loads in the optimization formulation are generated by one load spring per output DoF. In \cite{Zhan2010}, the ratio of a total mutual potential energy to the strain energy is maximized. The MPE and SE are calculated for each output DoF, and then combined in the objective function. Load springs act on the output DoF while computing the SE. This approach is tested on a mechanism with pseudo-mobility of two.

In some of the presented synthesis approaches, the synthesis additionally requires that "decoupled" mechanisms result \cite{Du2016,Hao2015,Tang2006,Wang2008,Zhan2010,Zhu2018}. According to \cite{Du2016}, decoupling can be divided into output decoupling and input decoupling. The base for the definition is that in each case an input DoF displacement causes a defined displacement in an associated output DoF. Output decoupling means that an input displacement only causes the prescribed displacement of the associated output DoF, but all other output DoFs remain unaffected. Input decoupling means that when one input DoF is displaced, the other input DoFs are not additionally displaced.

A first general limitation of state-of-the-art, optimization-based approaches is the way to cope with transverse loads. When they are considered, they enter the optimization formulation in combination with other load cases (no loads or loads at other outputs) and so they can be taken into account only in form of a weighted average, where the weights are arbitrary. For instance, when the entries of the stiffness matrix are used in the objective function, like in \cite{Pham2017}, the single transverse loads are weighted with the same coefficient. There, anyway, is mostly no request that the system is insensitive to transverse loads independently on the particular load case. A second limitation, which applies specifically to the multiple pseudo-mobility case, is that only simple kinematic specifications are treated. Deformation modes involving more than one DoF each are not considered up to now.

The so-called modal approach \cite{Hasse2009} distinguishes itself from other published structural-optimization based methods through its essential kinematic nature. It operates on the system’s eigenproperties, which – under certain conditions – describes preferred deformations independently on the acting loads. In this sense, and owing to its generality and applicability to systems with a large number of input and output DoF (collectively called “active DoF” – see below) \cite{Hasse2009,Hasse2016,Kirmse2021}, the modal approach represents a paradigm change in the area of compliant mechanism synthesis. It has the potential of covering a large part of the gaps left by the present state of the art.
Up to now, the modal approach was applied to design problems with single pseudo-mobility \cite{Hasse2009,Hasse2016,Kirmse2021}. In this paper, the extension to multiple pseudo-mobility is described and documented.

%% file: sec_3_0_Design_Problem.tex
\section{Design Problem}
\label{sec:Design Problem}
Owing to the essentially load-case independent nature of the modal approach, the distinction between Input- and Output-DoF falls away. Any DoF that can be subjected to loads and/or is involved in the definition of desired deformations is considered as active. The remaining DoFs are passive.
As explained above, compliant mechanisms have a theoretically infinite number of DoFs. For the following considerations, they are discretized to $p$ DoFs - for example by the finite element method. These are then divided into $q$ active and $p-q$ passive DoFs. Due to the fact that no external forces are allowed to act on the slave DoFs, the stiffness matrix \(\mathbf{K}\), the displacement vector \(\mathbf{u}\) and the force vector \(\mathbf{f}\) can be partitioned as follows: 
\begin{align}
    \label{Subdivision_stiffness_matrix}
    \mathbf{K} \mathbf{u} = \mathbf{f} \iff
    \begin{bmatrix}
        \mathbf{K}_{aa}&&\mathbf{K}_{ac}\\\mathbf{K}_{ca}&&\mathbf{K}_{cc}
    \end{bmatrix}
    \begin{bmatrix}
        \mathbf{u}_{a}\\\mathbf{u}_{c}
    \end{bmatrix}
    =
    \begin{bmatrix}
        \mathbf{f}_{a}\\\mathbf{0}
    \end{bmatrix}
\end{align}
Then, \(\mathbf{K}\) can be condensed to its active DoF's according to \cite{Gasch1989}:
\begin{align}
    \label{Condensation}
    (\mathbf{K}_{aa}-\mathbf{K}_{ac}\mathbf{K}_{cc}^{-1}\mathbf{K}_{ca})\mathbf{u}_a=\mathbf{\bar{K}}\mathbf{u}_{a}=\mathbf{f}_{a}
\end{align} 
In equations \eqref{Subdivision_stiffness_matrix} and \eqref{Condensation}, a proper support of the system is already considered, so that global rigid-body displacements are suppressed. All vectors $\mathbf{u}_{a} \in \mathbb{R}^q$ are then possible deformations of the mechanism. The vector space $\mathbb{R}^q$ is now subdivided into two subsets: the $m$-dimensional subspace of desired deformations and its complement. This is done by specifying a vector base for the subspace of desired deformations:
\begin{align}
    \label{desired_deformations}
    \bar{\bm{\Phi}}=
    \begin{bmatrix}
    \bar{\bm{\upvarphi}}_{1}&\bar{\bm{\upvarphi}}_{2}&\ldots&\bar{\bm{\upvarphi}}_{m}
    \end{bmatrix};
\end{align} 
The vectors $\bar{\bm{\upvarphi}}_{i}, i=1...m$ are the above introduced desired deformation modes. The eigenmodes \(\bm{\upchi}_j,j=1...q\) and the corresponding eigenvalues \(\lambda_j,j=1...q\) are determined for a compliant mechanism via the solution of the eigenvalue problem:
\begin{align}
    \label{Eigenvalue problem}
    \mathbf{\bar{K}}\bar{\bm{\mathrm{X}}}=\lambda\bar{\bm{\mathrm{X}}}
\end{align} 
As usual, the eigenvalues are sorted in ascending order and the eigenmodes correspondingly. The first $m$ eigenmodes are collected into the subset of \textit{kinematic eigenmodes} $\bar{\bm{\mathrm{X}}}_{d}$ and the remaining $q-m$ eigenmodes in the subset of the \textit{parasitic eigenmodes} $\bar{\bm{\mathrm{X}}}_{ud}$:
\begin{align}
    \label{Subdivision_eigenmodes}
    \bar{\bm{\mathrm{X}}}=
    \begin{bmatrix}
    \bar{\bm{\mathrm{X}}}_{d}&|&\bar{\bm{\mathrm{X}}}_{ud}
    \end{bmatrix}
\end{align} 
\begin{align}
    \label{Subdivision_eigenmodes2}
    \bar{\bm{\mathrm{X}}}_d=
    \begin{bmatrix}
    \bar{\bm{\upchi}}_{1}&\bar{\bm{\upchi}}_{2}&\ldots&\bar{\bm{\upchi}}_{m}
    \end{bmatrix};
    \bar{\bm{\mathrm{X}}}_{ud}=
    \begin{bmatrix}
    \bar{\bm{\upchi}}_{m+1}&\bar{\bm{\upchi}}_{m+2}&\ldots&\bar{\bm{\upchi}}_{q}
    \end{bmatrix};
\end{align} 
For the following, it is required that:
\begin{align}
    \label{Normalization}
    \bar{\bm{\upchi}}_i^T\bar{\bm{\chi}}_i=1, \; i=1...q
\end{align}
The eigenmodes \(\bm{\upchi}_j,j=1...q\) are, in addition to orthogonality with respect to the unit matrix $\mathbf{{{I}}}$, orthogonal with respect to the stiffness matrix, such that holds:
\begin{align}
    \label{Orthogonal}
    \left.
    \begin{matrix}
    \bar{\bm{\upchi}}_i^T\mathbf{\bar{K}}\bar{\bm{\chi}}_j=0 \\
    \bar{\bm{\upchi}}_i^T\mathbf{{I}}\bar{\bm{\chi}}_j=0
    \end{matrix} \right\} i=1...q, \; j=1...q
    \; ,i \not= j
\end{align}
With the conditions \eqref{Normalization}, the quantities
\begin{align}
    \label{Primaersteifigkeit}
    K_{pi}(\mathbf{\bar{K}})=\lambda_{i}=\bar{\bm{\upchi}}_{i}^T\mathbf{\bar{K}}\bar{\bm{\upchi}}_{i}, \; i=1\ldots m
\end{align}
can be called primary stiffnesses of the compliant mechanism \cite{Kirmse2021}. The secondary stiffness is calculated as:
\begin{align}
    \label{Secondary_Stiffness}
    K_s(\mathbf{\bar{K}})=\lambda_{\bar{\bm{\upchi}}_{m+1}}=\bar{\bm{\upchi}}_{m+1}^T\mathbf{\bar{K}}\bar{\bm{\upchi}}_{m+1}
\end{align}
The primary stiffnesses correspond to the eigenvalues \(\lambda_{1}\ldots\lambda_{m}\) associated to the kinematic eigenmodes. The secondary stiffness corresponds to the smallest eigenvalue \(\lambda_{m+1}\) associated to the parasitic eigenmodes. While in ideal conventional mechanisms the deformation response is a linear combination of $p$ desired modes, in compliant mechanisms it usually consists of a combination of kinematic and parasitic eigenmodes. The influence of the parasitic eigenmodes on the deformation response is increasingly reduced with an increasing value of the \textit{selectivity}:
\begin{align}
    \label{Selektivity}
    S=\lambda_{m+1}/\lambda_{m}
\end{align}
The higher the selectivity, the more accurately a deformation $\mathbf{\bar{u}}_d$ of the mechanism can be expressed as a linear combination of the kinematic eigenmodes:
\begin{align}
    \label{Desire_movement}
    \mathbf{\bar{u}}_d=\alpha_1\cdot \bar{\bm{\upchi}}_{1}+\alpha_2\cdot \bar{\bm{\upchi}}_{2}+\ldots+\alpha_m\cdot \bar{\bm{\upchi}}_{m}
\end{align}
The scaling factors $\alpha$ have the dimension of a length.

The main logics of the modal approach for the multiple pseudo-mobility case consists in requiring:
\begin{itemize}
\item that the subspace of desired deformations is spanned by the kinematic eigenmodes;
\item that the selectivity reaches a high value.
\end{itemize}

%% file: sec_4_0_Optimization_Formulation.tex
\section{Optimization formulation}
\label{sec:Optimization formulation}

\subsection{Overview}
\label{subsec:Optimization formulation}

As mentioned earlier, the optimization procedure described here is an adaptation of the procedure presented in \cite{Kirmse2021} to the more general case of compliant mechanisms with multiple pseudo-mobility. Therefore, the basic flow of the optimization and many of the formulas presented in \cite{Kirmse2021} retain their validity.

As mentioned, the first goal of the optimization is to determine the design variables $x_i,i=1... r$ such that the vector space $\mathbb{K}^{Xd}$ spanned by the first $m$ eigenmodes of the condensed stiffness matrix \(\mathbf{\bar{K}}\) resembles as closely as possible the space $\mathbb{K}^{\varphi}$ spanned by the desired deformation modes:
\begin{align}
    \label{Similarity}
    \mathbb{K}^{Xd} \approx \mathbb{K}^{\varphi}
\end{align}

Planar structures are assumed for the following considerations. Extension to non-planar mechanisms can be achieved by a proper adjustment of the parameterisation, i.e. of the functional relationship between stiffness matrix and design variables. In the present case, the ground structure method \cite{Bendsoe2003} was used for this, which suits the nature of compliant systems as an arrangement of flexural elements. Beams with combined axial and bending behavior are used here as flexural elements.

The variation of the stiffness as a function of the individual beam elements is done by a linear combination of the re-sized element stiffnesses \(\mathbf{K}_i\) with the design variables \(x_i\) as coefficients:
\begin{align}
    \label{Ground_structure_method}
    \mathbf{K}(\mathbf{x})= \underset{i=1}{\overset{r}{\sum}} x_i \mathbf{K}_i
\end{align}
The matrix \(\mathbf{K}(\mathbf{x})\) can then be reduced to \(\mathbf{\bar{K}}(\mathbf{x})\) using formula \eqref{Condensation}.

The desired deformation modes can be freely selected provided that following conditions are fulfilled:
\begin{align}
    \label{Conditions_u}
    \left.
    \begin{matrix}
     \bar{\bm{\upvarphi}}_i^T \bar{\bm{\upvarphi}}_j=1, \; i=j \\ \bar{\bm{\upvarphi}}_i^T \bar{\bm{\upvarphi}}_j=0, \; i \not= j
    \end{matrix} \right\}, i=1...m, \; j=1...m
\end{align}
These conditions ensure the normalization and orthogonality of the desired deformation modes.
To enforce orthogonality of a given, non-orthogonal set of deformation modes, a proper procedure, such the Gram-Schmidt's orthogonalization procedure (\cite{Kielbasinski1988}, pp. 277 to 285), can be used. 

The second goal of optimization, i.e. a high value of the selectivity $S$, can be indirectly achieved by maximizing the secondary stiffness $K_s$:
\begin{align}
    \label{Secondary_stiffness_general}
    \mathrm{max} \; f(\mathbf{x})=K_s(\mathbf{\bar{K}})
\end{align}
if at the same time all primary stiffnesses are subject to an upper limit. In the present work, for simplicity, all limits to the primary stiffnesses are set to the same value \(\mu\). Since the goal of the optimization algorithm involves that \eqref{Similarity} holds, the bounds on the primary stiffnesses are obtained by limiting the corresponding expression for the desired deformation modes
\begin{align}
    \label{Restriction_primary_stiffness_general}
    g_i(\mathbf{x})=\bm{\upvarphi}_i^T \mathbf{\bar{K}} \bm{\upvarphi}_i -\mu \le 0, i=1 \ldots m
\end{align}
In addition, a volume constraint, as usual for topology optimization, is introduced
\begin{align}
    \label{Volume_constraint_general}
    m(\mathbf{x})=\sum_{i=0}^r x_i - V \le 0
\end{align}
and the allowable range for the design variables is defined:
\begin{align}
    \label{Restriction_x_general}
    x_{i}^l \le x_{i} \le x_{i}^u, i=1...r
\end{align}
The lower bound in \eqref{Restriction_x_general} is used to prevent numerical problems and the upper bound is introduced for physical reasons.

The presented optimization problem \eqref{Secondary_stiffness_general}-\eqref{Restriction_x_general} is difficult to solve due to its nonlinearity. Therefore, it is divided into two subproblems, which renders a stepwise linear solution possible. This leads to a fast and efficient synthesis procedure, which also allows parameterizations with very large numbers of beam elements. In the first subproblem, the stiffness matrix is kept constant and an approximation of its eigenmodes (orthonormal base) is sought: 
\begin{align}
    \label{orthonormal_base_general}
    \mathbf{\bar{\Psi}}= [\bm{\bar{\upvarphi}}_1,\bm{\bar{\upvarphi}}_2,\ldots,\bm{\bar{\upvarphi}}_n,\bm{\bar{\uppsi}}_1,\bm{\bar{\uppsi}}_2,\ldots,\bm{\bar{\uppsi}}_n]
\end{align}
where the desired deformation modes $\bm{\bar{\upvarphi}}$ are fixed and the \textit{undesired modes} $\bm{\bar{\uppsi}}$ are variable. The orthonormal base is then expanded to all structural DoFs. This step is followed by a step of the second subprocedure, in which \(\mathbf{K}(\mathbf{x})\) is varied and improved by linear optimization. Both steps are performed alternately within a global iterative procedure, described in detail in the next subsections.

\subsection{Sub-problem 1: calculation of the orthonormal base and expansion}
\label{subsec:Sub-problem 1: calculation of the orthonormal base and expansion}

The variable part of the orthonormal base \(\mathbf{\bar{\Psi}}\) for a given \(\mathbf{\bar{K}}\) is calculated using the following optimization formulation:
\begin{align}
    \label{Golub}
    \mathrm{min} \; f(\bar{\bm{\uppsi}}_j)= \bar{\bm{\uppsi}}_j^T \mathbf{\bar{K}} \bar{\bm{\uppsi}}_j
\end{align}
s.t.:
\begin{align}
    \label{Golub_Con1}
    \begin{aligned}
      g_1(\bar{\bm{\uppsi}}_j) &= \bar{\bm{\upvarphi}}_1^T \mathbf{\bar{K}} \bar{\bm{\uppsi}}_j =0 \\ &\vdots \\ g_m(\bar{\bm{\uppsi}}_j) &= \bar{\bm{\upvarphi}}_m^T \mathbf{\bar{K}} \bar{\bm{\uppsi}}_j =0 \\
    \end{aligned} 
\end{align}
\begin{align*}
    \label{Golub_Con2}    
    g_{m+1}(\bar{\bm{\uppsi}}_j) &= \bar{\bm{\uppsi}}_1^T \mathbf{\bar{K}} \bar{\bm{\uppsi}}_j =0 ,\;j=2 \\
\end{align*}
\begin{align}
    \left. 
    \begin{aligned}
        g_{m+1}(\bar{\bm{\uppsi}}_j) &= \bar{\bm{\uppsi}}_1^T \mathbf{\bar{K}} \bar{\bm{\uppsi}}_j =0 \\ &\vdots \\ g_{m+j-1}(\bar{\bm{\uppsi}}_j) &= \bar{\bm{\uppsi}}_{j-1}^T \mathbf{\bar{K}} \bar{\bm{\uppsi}}_j =0 \\
    \end{aligned} 
    \right\} j>2
\end{align}
\begin{align}
    \label{Golub_Con3}
    h(\bar{\bm{\uppsi}}_j) = \bar{\bm{\uppsi}}_j^T \bar{\bm{\uppsi}}_j =1
\end{align}
This optimization formulation is to be solved recursively for $\bar{\bm{\uppsi}}_j$ for increasing $j$ from 1 to $q-m$. This corresponds to the restricted search for the extreme values of the quadratic form on the right-hand side of \eqref{Golub} with the constraints \eqref{Golub_Con1} and \eqref{Golub_Con3}. A computationally efficient solution to this problem is described in \cite{Golub1973b}. With this approach, all $q-m$ vectors of the orthonormal base are computed using a substitute eigenvalue problem. For the correct positioning in $\mathbf{\bar{\Psi}}$, the vectors of the orthonormal base are then to be ordered according to the function values computed according to \eqref{Golub} in ascending order. Then, all vectors of $\mathbf{\bar{\Psi}}$ are expanded to all structural DoF's as follows:
\begin{align}
    \label{nr18}
    \bm{\upvarphi}_i=
    \begin{bmatrix}
        \bar{\bm{\upvarphi}}_i \\ -\mathbf{K}_{aa}^{-1} \mathbf{K}_{ca} \bar{\bm{\upvarphi}}_i
    \end{bmatrix}
    ; \;
    \bm{\uppsi}_j=
    \begin{bmatrix}
        \bar{\bm{\uppsi}}_j \\ -\mathbf{K}_{aa}^{-1} \mathbf{K}_{ca} \bar{\bm{\uppsi}}_j
    \end{bmatrix}
\end{align}

\subsection{Sub-problem 2: updating the design variables}
\label{subsec:Sub-problem 2: updating the design variables}

The entries in the orthonormal base $\mathbf{\bar{\Psi}}$ are used to solve the second sub-problem, where the maximization of the objective function \eqref{Secondary_stiffness_general} with the constraints \eqref{Restriction_primary_stiffness_general}, \eqref{Volume_constraint_general} and \eqref{Restriction_x_general} is performed. 
The second sub-problem is as follows:
\begin{align}
    \label{Maximization_secondary_stiffness}
    \mathrm{max} \; f(\mathbf{x})=\bm{\uppsi}_1^T \mathbf{K}(\mathbf{x}) \bm{\uppsi}_1
\end{align}
s.t.:
\begin{align}
    \label{Restriction_primary_stiffness}
    \begin{matrix}
        g_1(\mathbf{x})=\bm{\upvarphi}_1^T \mathbf{K} \bm{\upvarphi}_1 -\mu \le 0 \\ \vdots \\
        g_m(\mathbf{x})=\bm{\upvarphi}_m^T \mathbf{K} \bm{\upvarphi}_m -\mu \le 0
    \end{matrix}
\end{align}
\begin{align}
    \label{Orthogonality_modes_to_K}
    h_1(\mathbf{x}) \ldots h_{\binom{m}{2}}(\mathbf{x}) =\bm{\upvarphi}_i^T \mathbf{K} \bm{\upvarphi}_j=0, \; i \not= j, \; i=1 \ldots m, \; j=1 \ldots m
\end{align}
\begin{align}
    \label{Restriction_mode_swap}
    \begin{matrix}
    k_1(\mathbf{x})=\bm{\uppsi}_1^T \mathbf{K}(\mathbf{x}) \bm{\uppsi}_1-\bm{\uppsi}_2^T \mathbf{K}(\mathbf{x}) \bm{\uppsi}_2\le 0 \\ \vdots \\
    k_n(\mathbf{x})=\bm{\uppsi}_1^T \mathbf{K}(\mathbf{x}) \bm{\uppsi}_1-\bm{\uppsi}_n^T \mathbf{K}(\mathbf{x}) \bm{\uppsi}_n\le 0 
    \end{matrix}
\end{align}
\begin{align}
    \label{Volume_restriction}
   m(\mathbf{x})=\sum_{i=0}^r x_i - V \le 0
\end{align}
\begin{align}
    \label{Restriction_boundaries_global}
    x_{i}^l \le x_{i} \le x_{i}^u, \; i=1...r
\end{align}
The minimization \eqref{Maximization_secondary_stiffness} corresponds to the equation \eqref{Secondary_stiffness_general}, as mentioned. The constraint \eqref{Restriction_primary_stiffness} corresponds to the constraint \eqref{Restriction_primary_stiffness_general}, rewritten for the uncondensed system. The constraints \eqref{Volume_restriction} and \eqref{Restriction_boundaries_global} coincide with \eqref{Volume_constraint_general} and \eqref{Restriction_x_general}. The conditions \eqref{Orthogonality_modes_to_K} enforce the orthogonality of the desired deformation modes $\bm{\upvarphi}$ with respect to \(\mathbf{K}\), which strongly improves the quality of the solutions. Without this restriction, structures are also found which satisfy equation \eqref{Similarity} to a good degree, but with the first $m$ eigenvalues not similar to each other and with low selectivity. The conditions \eqref{Restriction_mode_swap} ensure that the vector which should approximate the $m+1$. eigenmode of the stiffness matrix (ordered by increasing eigenvalue) does not swap its rank with another deformation mode, as this could destabilize the procedure. All $q-m$ undesirable deformation modes can be included in the optimization. In systems with many active DoFs, this can lead to an unconvenient increase of the computational cost. In this case, a smaller number of undesired modes can be used as a compromise between stability and low computational cost. 

Due to its linearity, this subproblem can be solved very efficiently. This is achieved by the dual simplex algorithm \cite{Nelder1965}.

In contrast to the single pseudo-mobility case, the synthesis of selective compliant mechanisms with multiple pseudo-mobility can lead to solutions in which the first $m$ eigenmodes do not approximate the desired deformation modes. This is not of relevance, as long as the first $m$ eigenmodes span the same space as $\mathbb{K}^{xd}$. If the same limit $\mu$ is set for the primary stiffness of all desired deformation modes, the solutions tend to an eigenvalue with geometric multiplicity $m$. The space $\mathbb{K}^{xd}$ can then be described by different sets of $m$ linearly independent vectors, which are all eigenmodes of the mechanism's stiffness matrix.

\subsection{Global iteration procedure}
\label{subsec:Global iteration procedure}

It has already been mentioned that subproblems 1 and 2 must be solved step by step alternately in an iterative manner. The output of subproblem 1, i.e., the extended orthonormal base, serves as the input for subproblem 2. The computed set of design variables from subproblem 2, in turn, represents the input for subproblem 1. At convergence, the final solution is found for the vector of design variables $\mathbf{x}$. In the following, two consecutive steps of subproblems 1 and 2 are referred to as iteration step. 

In order to stabilize the global iteration procedure, the variation of the design variables between two iteration steps must be restricted. Therefore, a restriction is added to the optimization problem, which acts in addition to restriction \eqref{Restriction_boundaries_global}. The additionally inserted lower and upper bounds restrict the possible range of the design variables and thus the possible change of $\mathbf{K}$. The bounds for the next iteration step are set up based on the currently calculated values for $\mathbf{x}$ as follows:
\begin{align}
    \label{nr24}
    \begin{matrix}
    \mathbf{x}_t^l (s+1)= \mathbf{x}_t(s)+\bm{\upnu} \\
    \mathbf{x}_t^u (s+1)= \mathbf{x}_t(s)-\bm{\upnu}
    \end{matrix}
\end{align}

The variable $s$ defines the current iteration step and $\bm{\upnu}$ is a process parameter. The possible range of the design variables should be wide for the first iteration steps, so that the algorithm better orientates toward possible solutions of the problem. Then it should be reduced. How large the elements of $\upnu$ can be chosen depends on the one hand on the number of design variables and on the other hand on the number of stabilizing deformation modes. If the limits per step are too large and no restriction \eqref{Restriction_mode_swap} is used, the optimization algorithm may not converge to a useful result because the change of the stiffness matrix between both subproblems may become too large.

\begin{figure*}[t]
    \centering
    \includegraphics[width=16cm]{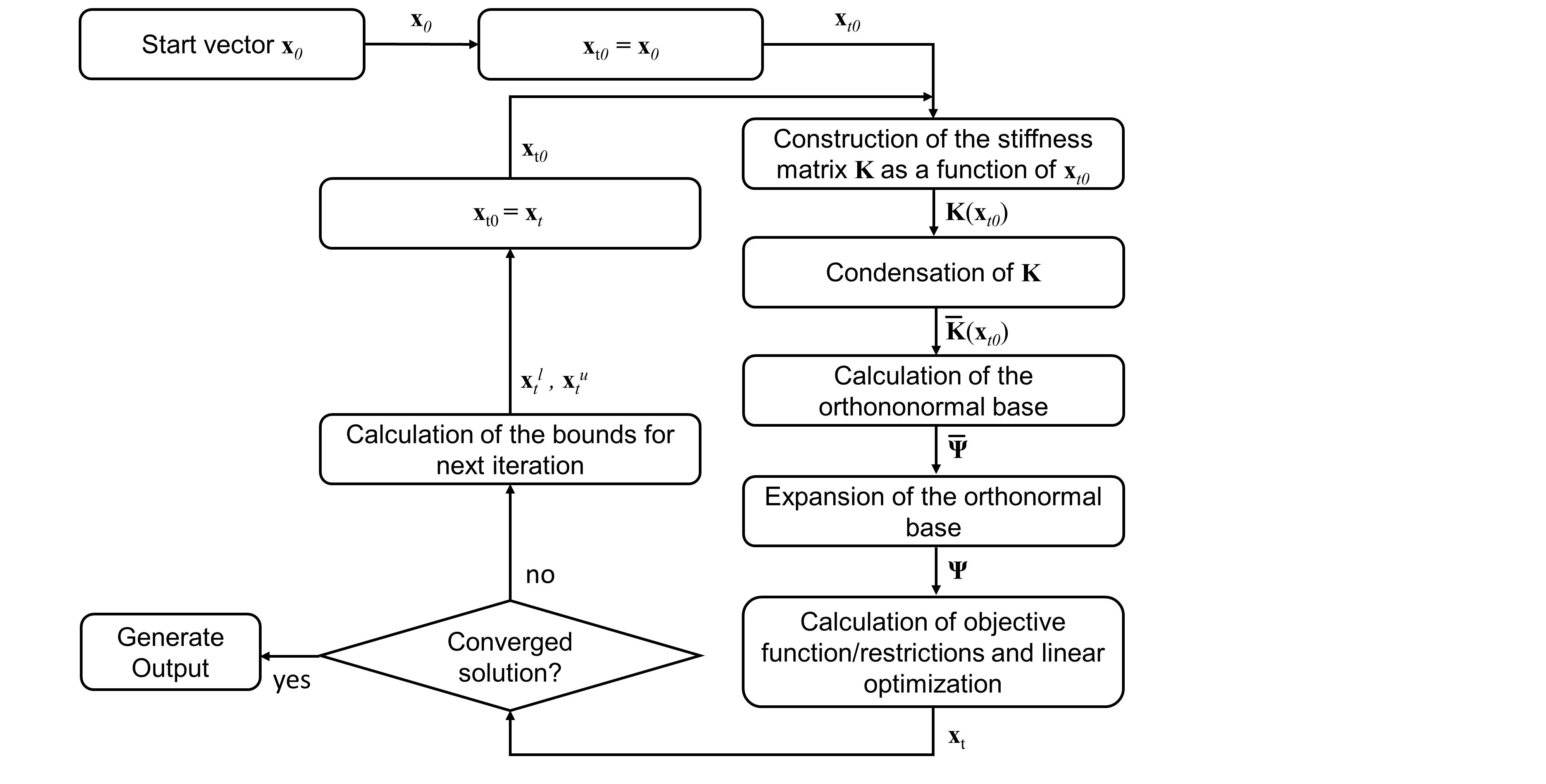}
    \caption{Global iteration procedure}
    \label{figure1}
\end{figure*}

For the optimization procedure, a starting vector $\mathbf{x}_0$ must be chosen. Depending on this, the optimization can converge to different solutions since the global problem is non-convex. The procedure is illustrated in \autoref{figure1}.

%% file: sec_5_0_Design_Examples.tex
\section{Design examples}
\label{sec:Design examples}

The presented optimization algorithm was implemented in MATLAB and tested for different design examples. Two mechanisms with simple kinematics as well as a shape-adaptive structure serve as design examples. The first design example is a mechanism with a combination of a rotation and straight translation as desired kinematics, corresponding to a pseudo-mobility of two. The number of active DoFs is very small here. The second example is a compliant parallel mechanism with a pseudo-mobility of two. Here, a larger number of active DoFs was chosen for the synthesis, and the desired kinematics describes translational rigid-body displacements of the platform. The last example is a shape-adaptive structure with a pseudo-mobility of three, which allows a combination of a sinusoidal deformation, a deformation in the form of a parabola and a translational displacement of its contour. Here the number of active DoFs is very large. For all examples, beam elements with the following material and geometry parameters were chosen for parameterization:
\begin{align}
   A=20 \, \mathrm{mm^2}, \;  E=210 \, \mathrm{GPa}, \; I=6.66 \, \mathrm{mm^4} 
\end{align}
where $A$ is the cross-sectional area, $E$ the elastic modulus, and $I$ the area moment of inertia. 

In the implementation of the procedure discussed in this work, all design variables were bound by the same limits according to \eqref{Restriction_boundaries_global}:
\begin{align}
    \label{Restriction_limits_entry_lower}
    x^l_i=x^l, \; i=1...r
\end{align}
\begin{align}
    \label{Restriction_limits_entry_upper}
    x^u_i=x^u, \; i=1...r
\end{align}
In all examples $x_l=1 \cdot 10^{-8}$ and $x_l=1$ were chosen. 
Similarly, only vectors $\bm{\upnu}$ with the same value for all entries are considered:
\begin{align}
    \label{nu-entries}
    \upnu_i=\upnu ,\;i=1...r
\end{align}
In all examples $\upnu$ was chosen equal to 0.001.

Since, as mentioned, the solution depends on the starting value, it is necessary to perform several calculations starting from different vectors $\mathbf{x}_0$ to obtain the best possible solution. Uniformly distributed random numbers between $1 \cdot 10^{-8}$ and 1 were chosen as entries in $\mathbf{x}_0$. In addition, $\mu$ should be varied, as the achieved selectivity may vary depending on it. A detailed consideration of the influence of $\mu$ on the optimization result can be found in \cite{Kirmse2021}. The quality of the solutions is assessed according to two criteria: the level of selectivity achieved and the similarity of $\mathbb{K}^{xd}$ to $\mathbb{K}^{\varphi}$. In each of the following sections, solutions with high selectivity are shown. In order to evaluate the similarity of the two vector spaces to each other, the best approximation - in the least squares sense - of a desired deformation mode $\bm{{\upvarphi}}'_i$ by a linear combination of the kinematic eigenmodes can be determined:
\begin{align}
    \label{Approximation}
    \bm{\bar{\upvarphi}}'_i=\alpha_{i1} \cdot \bm{\bar{\upchi}}_1+\alpha_{i2} \cdot \bm{\bar{\upchi}}_2+ \ldots + \alpha_{im} \cdot \bm{\bar{\upchi}}_m \; i=1 \ldots m
\end{align}
and compared with $\bm{\bar{\upvarphi}}_i$. A quantitative comparison criterion is the extended cosine similarity presented in \cite{Campanile2021}, which provides one single scalar value as a similarity measure for the whole subspaces. For this, the following eigenvalue problem has to be solved first:
\begin{align}
    \label{Eigenvalue_cosine}
    \bm{\bar{\Phi}}^T \bm{\bar{\mathrm{X}}}_d \bm{\bar{\mathrm{X}}}_d^T \bm{\bar{\Phi}} \mathrm{\bm{b}} =\upbeta \mathrm{\bm{b}} 
\end{align}
Then, the extended cosine similarity is calculated as follows using the smallest eigenvalue $\upbeta_1$:
\begin{align}
    \label{Cosine_expanded}
    \updelta_e=\sqrt{\upbeta_1}
\end{align}

The extended cosine similarity produces values between zero and one, with higher values for increasing similarity. It reaches the value of one if the compared subspaces coincide. If two one-dimensional spaces are compared, it supplies the cosine similarity \cite{Singhal2001} of the two basis vectors.

\subsection{Mechanism with a combination of a rotation and a straight translation with double pseudo-mobility}
\label{simple_mechanism_with_douple_pseudo_running}

\begin{figure*}[t]
    \centering
    \includegraphics[width=16cm]{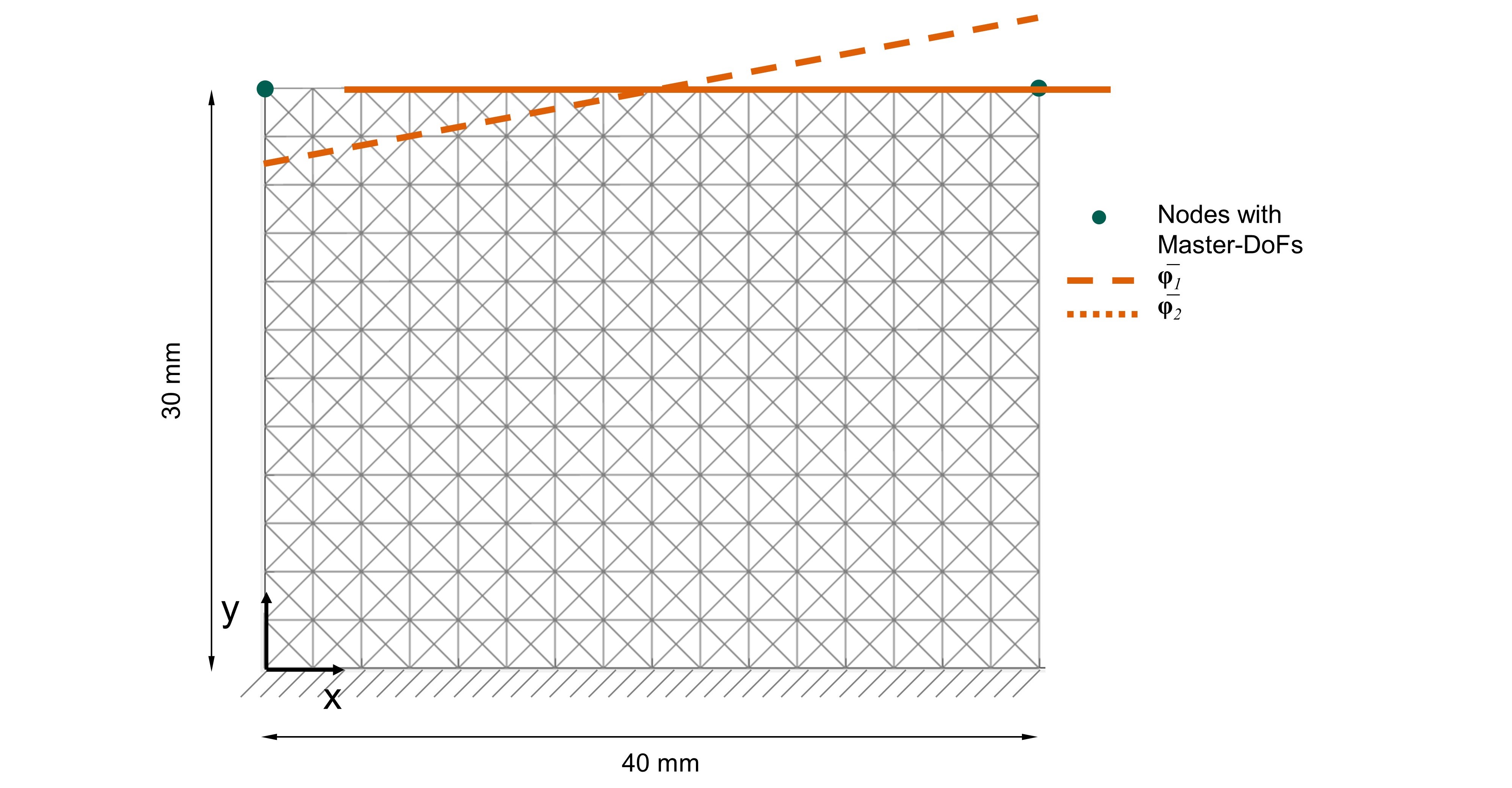}
    \caption{Double pseudo-mobility mechanism with a combination of a rotation and a straight translation: Parameterized design space with desired deformation modes $\bm{\bar{\upvarphi}}_i$}
    \label{figure2}
\end{figure*}

The design space chosen for this example and its dimensions can be seen in \autoref{figure2}. Here, 796 beam elements were used for parameterization. In total, the structure has 663 structural DoFs and was clamped at its lower side. Four active DoFs were defined, given by $x$- and $y$- displacements of the two points marked in \autoref{figure2}.
The first of the two chosen deformation modes $\bm{\bar{\upvarphi}}_1$ describes a rotation. Thereby the two selected points shall move vertically by the same amount in opposite directions. The second desired deformation mode $\bm{\bar{\upvarphi}}_2$ describes a straight translation. The two selected points move horizontally by the same amount in the same direction.
\begin{align}
    \label{phi1}
    \bm{\bar{\upvarphi}}_1=
    \begin{bmatrix}
        \bar{\upvarphi}_{11x} \\ \bar{\upvarphi}_{11y} \\ \bar{\upvarphi}_{12x} \\ \bar{\upvarphi}_{12y}
    \end{bmatrix}
    =
    \begin{bmatrix}
        0.707 \\ 0 \\ 0.707 \\ 0
    \end{bmatrix} 
    \\
    \label{phi2}
    \bm{\bar{\upvarphi}}_2=
    \begin{bmatrix}
        \bar{\upvarphi}_{21x} \\ \bar{\upvarphi}_{21y} \\ \bar{\upvarphi}_{22x} \\ \bar{\upvarphi}_{22y}
    \end{bmatrix}
    =
    \begin{bmatrix}
        0 \\ -0.707 \\ 0 \\ 0.707
    \end{bmatrix}        
\end{align}
The difference between the number of active DoF (four) and the pseudomobility (two) provides the number of undesired modes (two). $V$ is equal to 636.8, which corresponds to 80 \% of the maximum allowable volume. Seven uniformly distributed values of $\mu$ between 1000 and 4000 were chosen and 100 calculations each were performed. The condition \eqref{Restriction_mode_swap} was applied for both undesired modes.

\begin{figure*}[t]
    \centering
    \includegraphics[width=16cm]{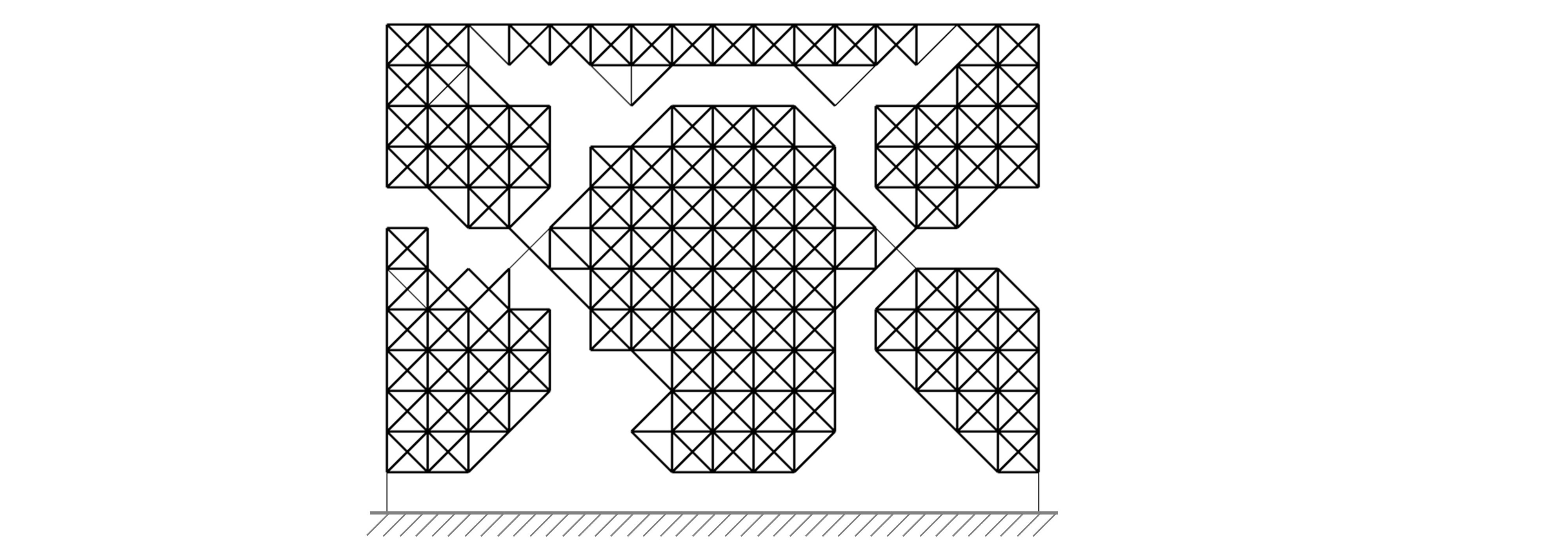}
    \caption{Double pseudo-mobility mechanism with a combination of a rotation and a straight translation: Solution for $\nu$ = 3000}
    \label{figure3}
\end{figure*}

\begin{figure*}[t]
    \centering
    \includegraphics[width=16cm]{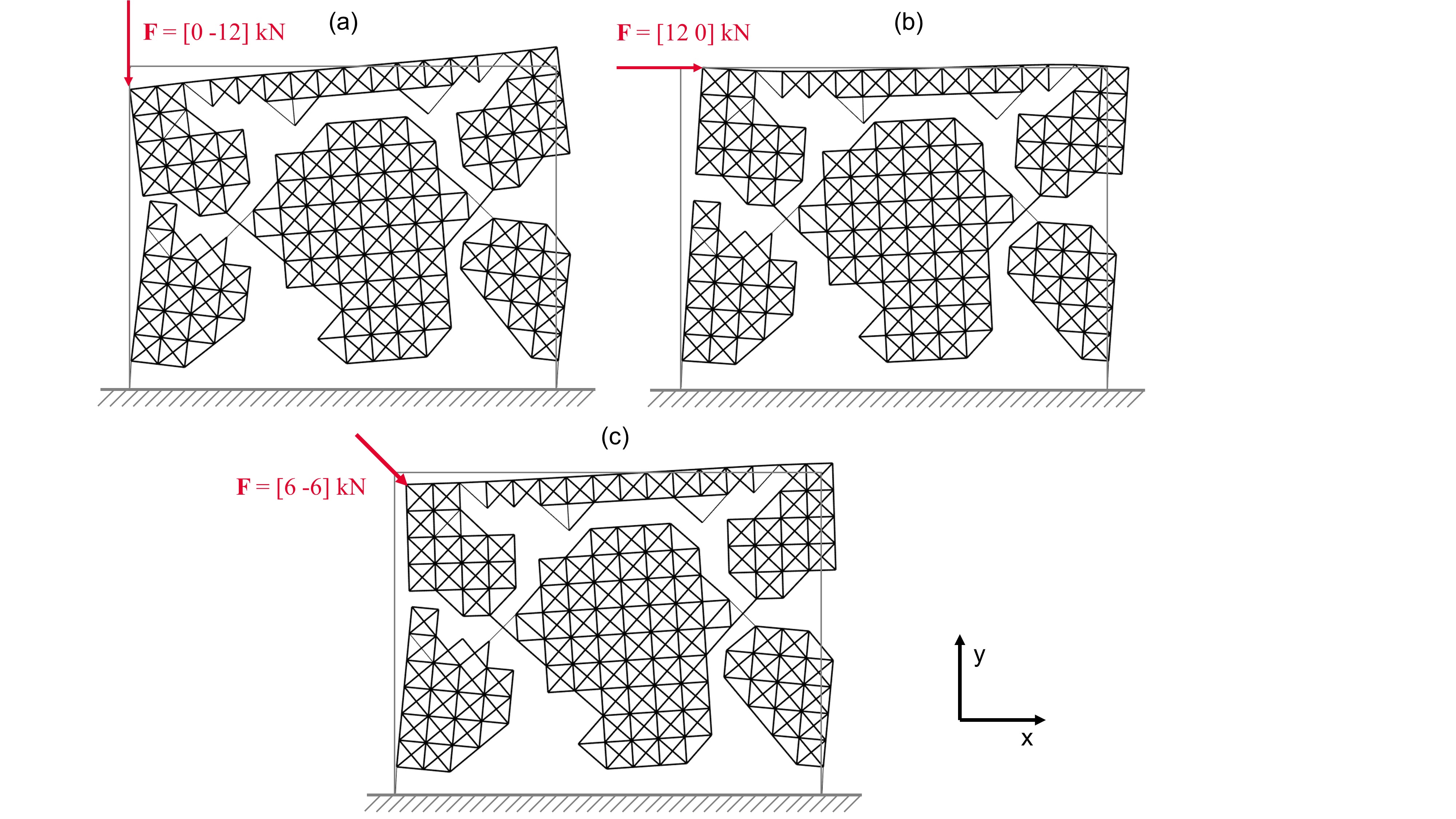}
    \caption{Double pseudo-mobility mechanism with a combination of a rotation and a straight translation: Different load cases}
    \label{figure4}
\end{figure*}

One of the resulting designs can be seen in \autoref{figure3}.
The eigenmodes of the structure and the corresponding eigenvalues can be seen in \autoref{Table1}.
\begin{table}
    \begin{center}
    \caption{Mechanism with a combination of a rotation and a straight translation: eigenvalue analysis}
    \begin{tabularx}{240pt}{@{}c|cccc@{}}
        \toprule
        $i=$ & 1 & 2 & 3 & 4 \\
        \midrule
        $\lambda_i$ & 2999.8 & 2999.9 & 80973.6 & 368667.9 \\    
        $\bm{\bar{\upchi}}_i$ & 
        $\begin{bmatrix}
            0.565 \\ 0.426 \\ 0.564 \\ -0.425
        \end{bmatrix}  $ &
        $\begin{bmatrix}
            -0.426 \\ 0.565 \\ -0.425 \\ -0.565
        \end{bmatrix}  $ &
        $\begin{bmatrix}
            -0.132 \\ -0.695 \\ 0.132 \\ -0.695
        \end{bmatrix}  $ &
        $\begin{bmatrix}
            0.694 \\ -0.132 \\ -0.695 \\ -0.132
        \end{bmatrix}  $ \\
        \bottomrule
    \end{tabularx}
    \label{Table1}
    \end{center}
\end{table}
The results show a large difference between the second and third eigenvalue. As described in \autoref{sec:Introduction}, this confirms the value of two for the pseudo-mobility. The selectivity is 27.0 and the first two eigenvalues are very similar. According to \eqref{Approximation}, the best least square approximation for the first desired mode is:
\begin{align}
    \bm{\bar{\upvarphi}}'_1= 0.80 \cdot \bm{\bar{\upchi}}_{1} - 0.60 \cdot \bm{\bar{\upchi}}_{2}
    =
    \begin{bmatrix}
        0.7076 \\ 0.0001 \\ 0.7066 \\ 0.0001
    \end{bmatrix}       
\end{align}
and for the second desired mode:
\begin{align}
    \bm{\bar{\upvarphi}}'_2= -0.60 \cdot \bm{\bar{\upchi}}_{1} - 0.80 \cdot \bm{\bar{\upchi}}_{2}
    =
    \begin{bmatrix}
        -0.0002 \\ -0.7072 \\ 0.0002 \\ 0.7070
    \end{bmatrix}       
\end{align}
Comparing with \eqref{phi1} and \eqref{phi2} show that the similarity of the vector spaces $\mathbb{K}^{xd}$ and $\mathbb{K}^{\varphi}$ is very high. The extended cosine similarity according to \autoref{Cosine_expanded} with 0.9999997 confirms this. In \autoref{figure4} the deformed structure can be seen under different load cases.
Both of the desired deformation modes can be addressed individually (\autoref{figure4} (a)-(b)), as well as a linear combination (\autoref{figure4} (c)).

\subsection{Parallel mechanism with double pseudo-mobility}
\label{Parallel mechanism with double pseudo-mobility}

\begin{figure*}[t!]
    \centering
    \includegraphics[width=16cm]{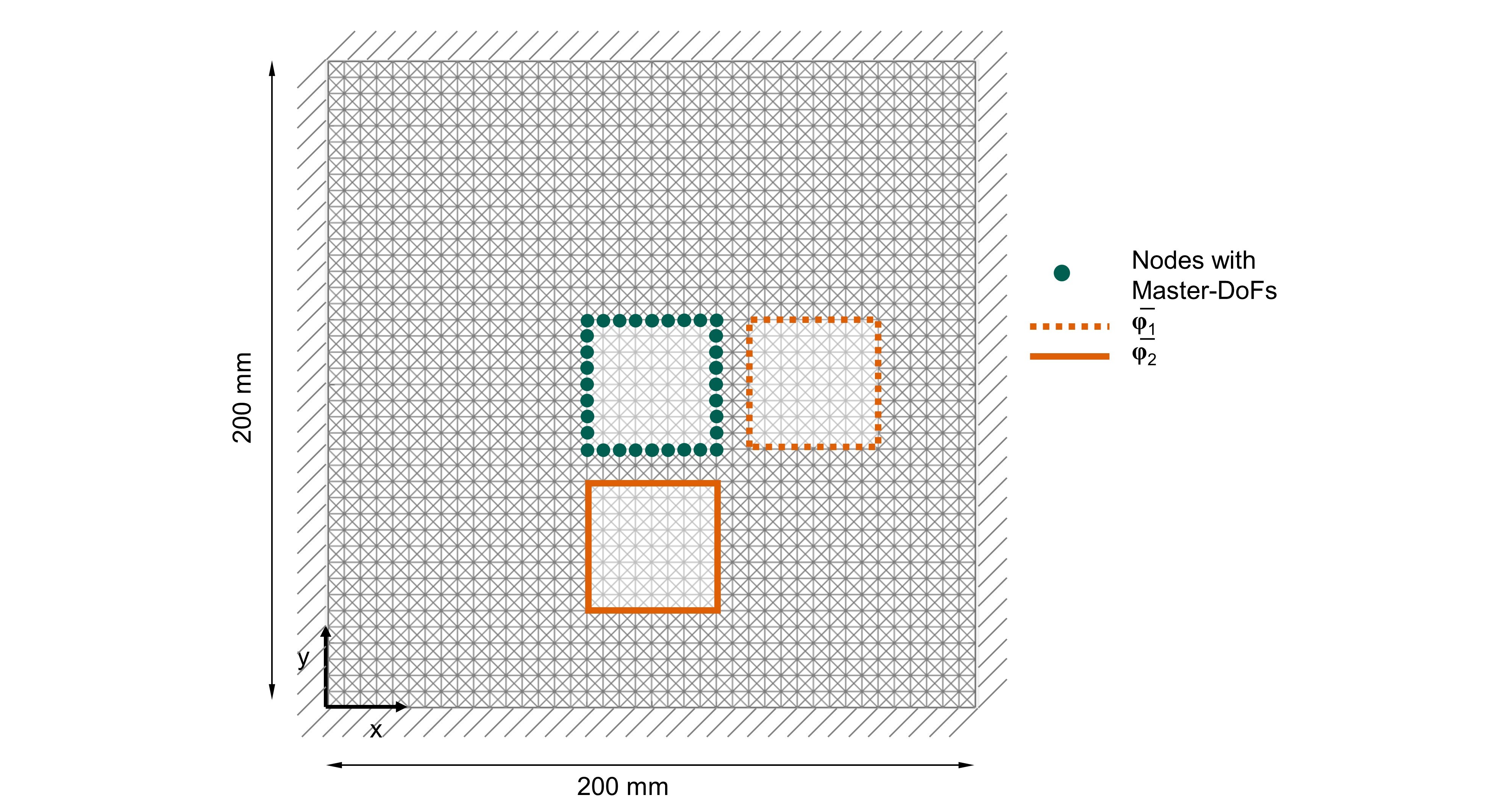}
    \caption{Compliant parallel mechanism: Parameterized design space and desired deformation modes $\bm{\bar{\upvarphi}}_i$}
    \label{figure5}
\end{figure*}

\begin{figure*}[t!]
    \centering
    \includegraphics[width=16cm]{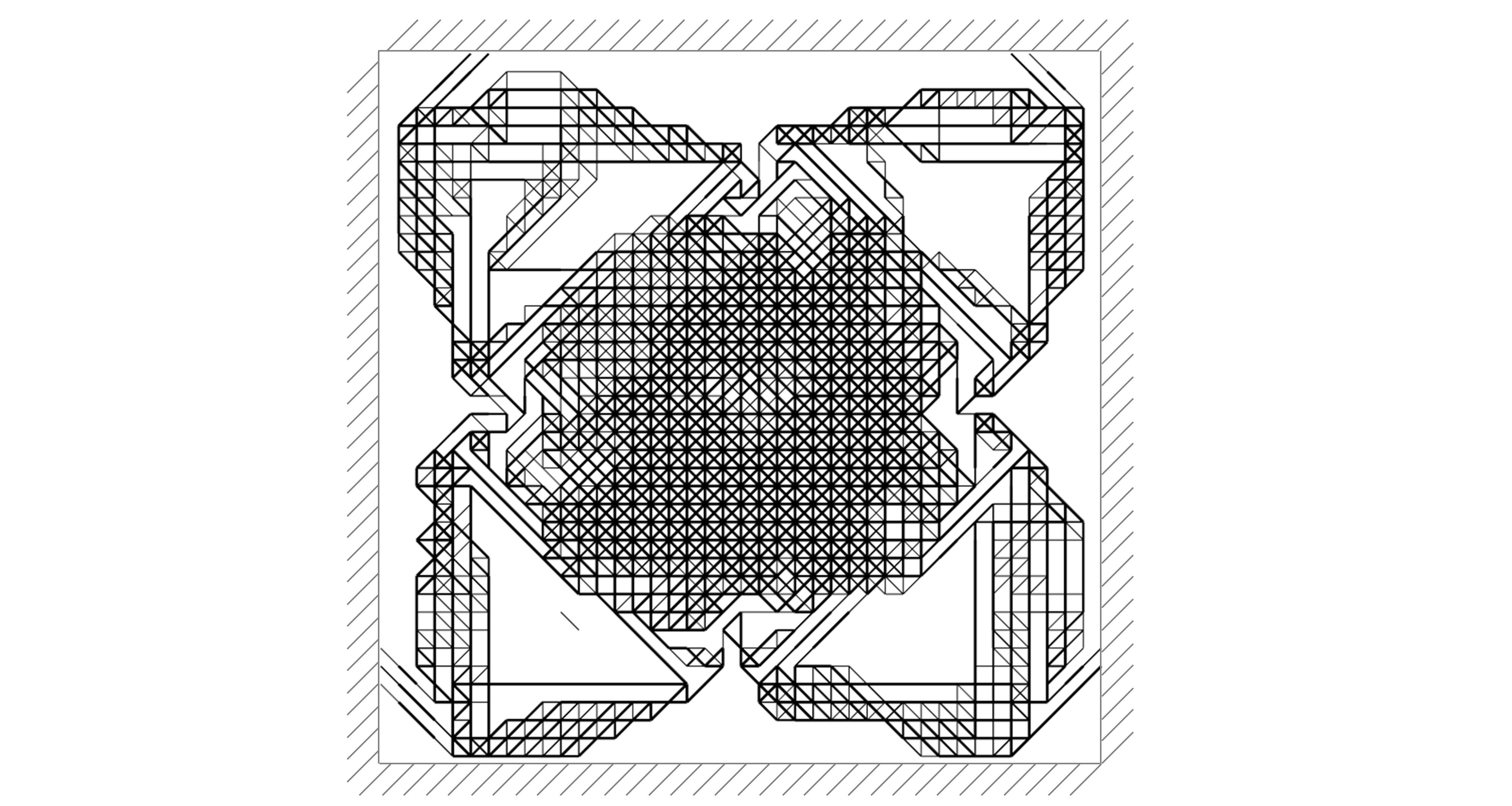}
    \caption{Compliant parallel mechanism: Solution for $\mu=500$}
    \label{figure6}
\end{figure*}

\begin{figure*}[t!]
    \centering
    \includegraphics[width=16cm]{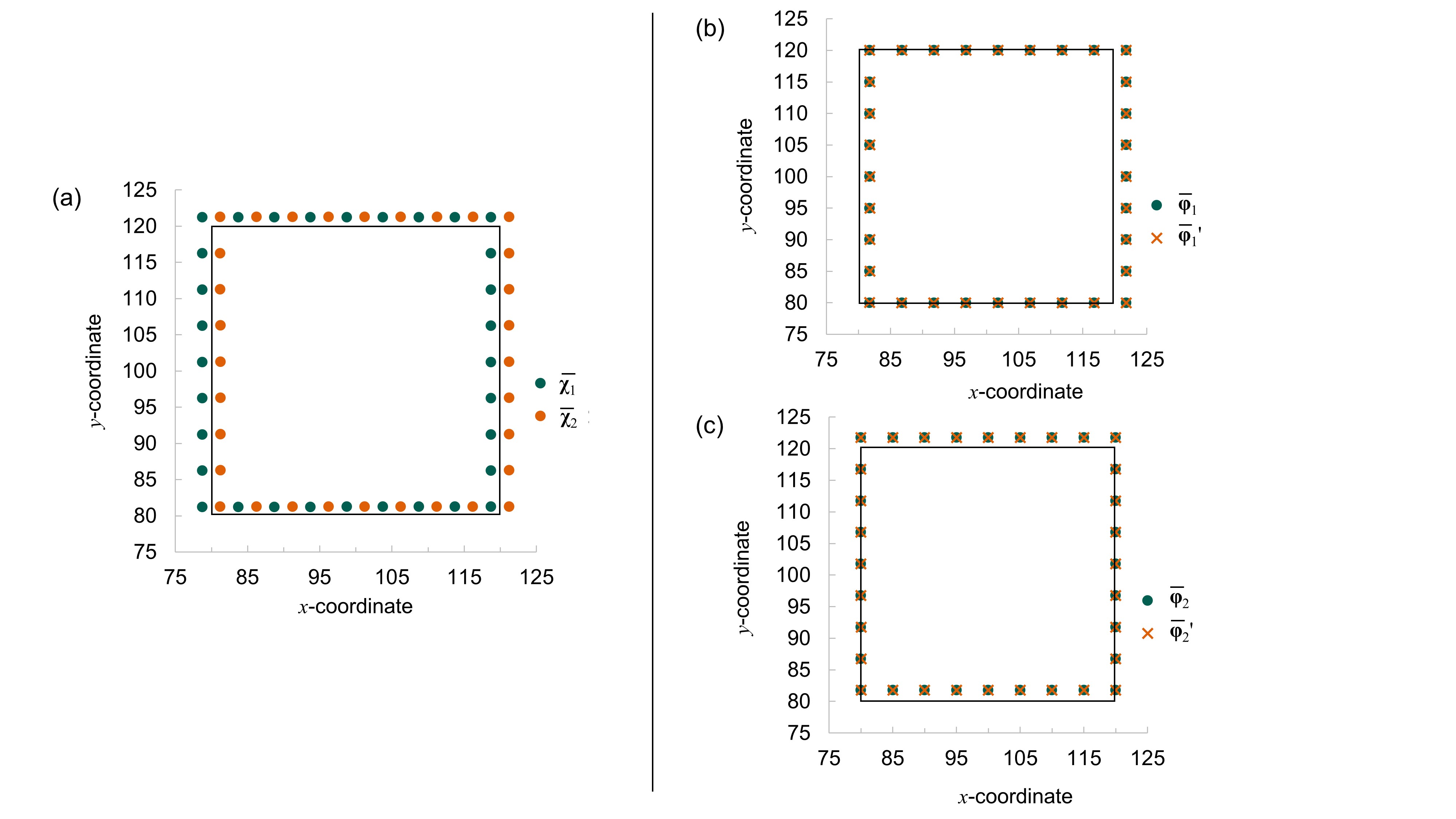}
    \caption{Compliant parallel mechanism: (a) Kinematic eigenmodes $\bm{\bar{\upchi}}_{i}$, (b)-(c) comparison of the best linear combination of the kinematic eigenmodes with the desired deformation modes $\bm{\bar{\upvarphi}}'_i$}
    \label{figure7}
\end{figure*}

The synthesis of compliant parallel mechanisms is a popular topic in the literature. Therefore, this case was also chosen here to show the suitability of the presented optimization algorithm for such mechanisms. The selected design space can be seen in \autoref{figure5}. The mechanism was fixed at all four boundaries. By a suitable choice of nodes with active DoFs, a rigid platform was defined in the center of the design space, which is required to be able to perform any translation, but no rotation. For this purpose, two desired deformation modes ($\bm{\bar{\upvarphi}}_1$, $\bm{\bar{\upvarphi}}_2$) were specified, which include all $x$- and $y$- displacements of the points bounding the platform. This results in a total of 64 active DoFs. The first desired deformation mode describes the horizontal motion of the platform. This means that the entries of the $x$-displacements of the nodes all have the same value, while the $y$-displacements vanish. The second desired deformation mode describes a vertical motion of the platform. Here, all $y$-displacements have the same value, while the $x$-displacements are zero.
The design space was parameterized with 6480 beam elements and 5043 structural DoFs. Here, 20 undesired modes were used to stabilize the mechanism according to equation \eqref{Restriction_mode_swap}. The allowed volume $V$ was set to the value 2592, which corresponds to 40 \% of the maximum allowable volume. Five calculations each were performed with four different $\mu$ equally spaced between 500 and 2000.
The selected result can be seen in \autoref{figure6}.
The eigenvalues belonging to the first five eigenmodes can be seen in \autoref{Table2}. Again, it can be seen that the first two eigenvalues are very similar. All other eigenvalues are much larger, and the selectivity amounts to 108.4.
\begin{table}
    \begin{center}
    \caption{Compliant parallel mechanism: Eigenvalue analysis}
    \begin{tabularx}{230pt}{@{}c|cccc|c@{}}
        \toprule
        $i=$ & 1 & 2 & 3 & 4 & S \\
        \midrule
        $\lambda_i$ & 991.2 & 993.0 & 107688.2 & 223390.0 & 108.4 \\
        \bottomrule
    \end{tabularx}
    \label{Table2}
    \end{center}
\end{table}
The first two eigenmodes are shown graphically in \autoref{figure7} (a). 
Again, it can be seen that they do not correspond to the desired deformation modes. However, also in this case proper similarity can be reached by linear recombination:
\begin{align}
    \bm{\bar{\upvarphi}}'_1= 0.72 \cdot \bm{\bar{\upchi}}_{1} - 0.70 \cdot \bm{\bar{\upchi}}_{2} 
\end{align}
\begin{align}
    \bm{\bar{\upvarphi}}'_2= -0.70 \cdot \bm{\bar{\upchi}}_{1} - 0.72 \cdot \bm{\bar{\upchi}}_{2}     
\end{align}
The similarity between $\bm{\bar{\upvarphi}}'_i$ and $\bm{\bar{\upvarphi}}_i$ can be seen in \autoref{figure7} (b) on the right. The vectors agree well, as also shown by the extended cosine similarity of 0.999993.

\begin{figure*}[t!]
    \centering
    \includegraphics[width=16cm]{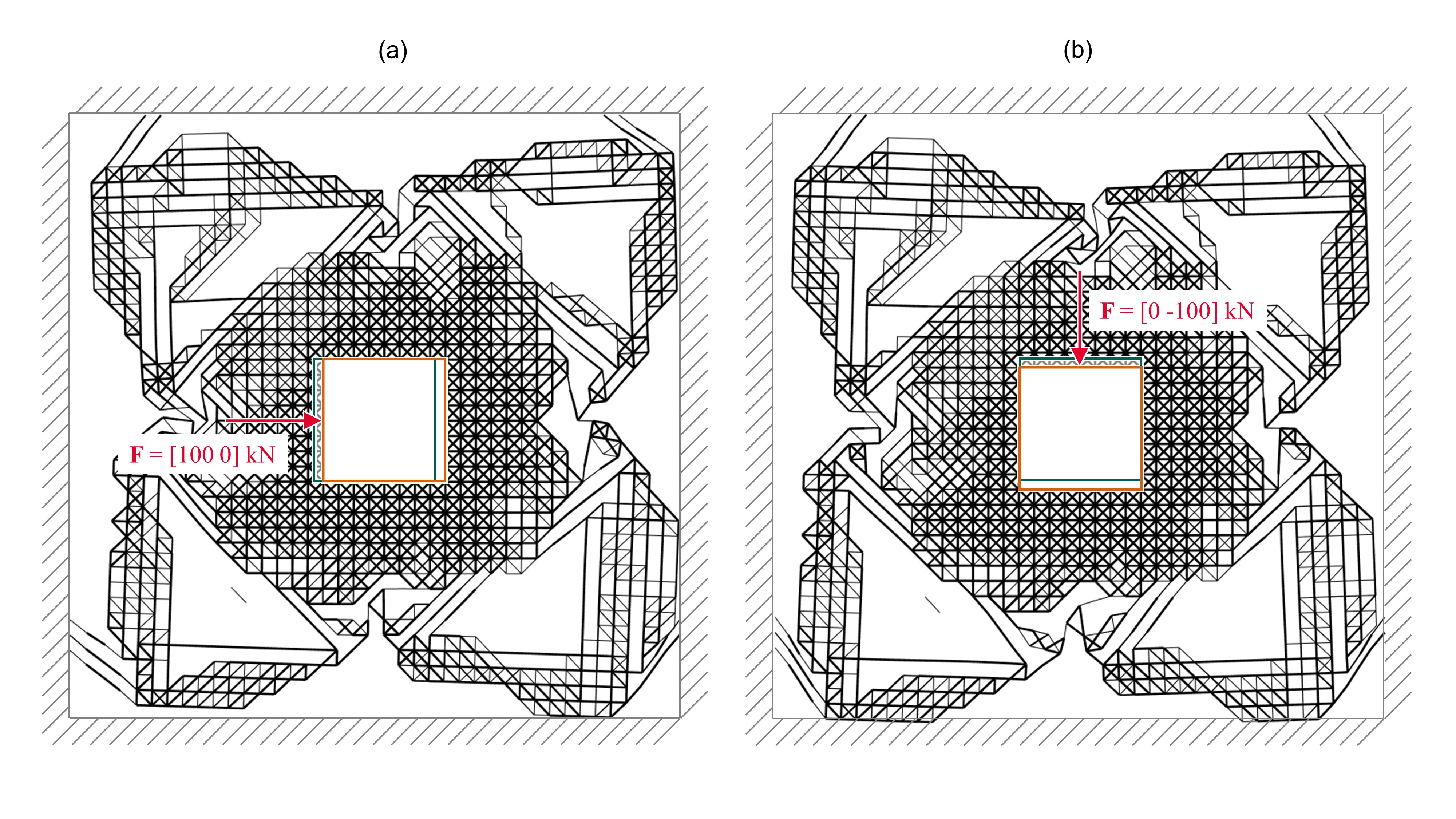}
    \caption{Compliant parallel mechanism: Different load cases}
    \label{figure8}
\end{figure*}

\autoref{figure8} shows the mechanism under different loads. The platform defined by the active DoFs is represented by a white square.

\subsection{Shape-adaptive structure with triple pseudo-mobility}
\label{Shape-adaptive structure with triple pseudo-mobility}

\begin{figure*}[t!]
    \centering
    \includegraphics[width=16cm]{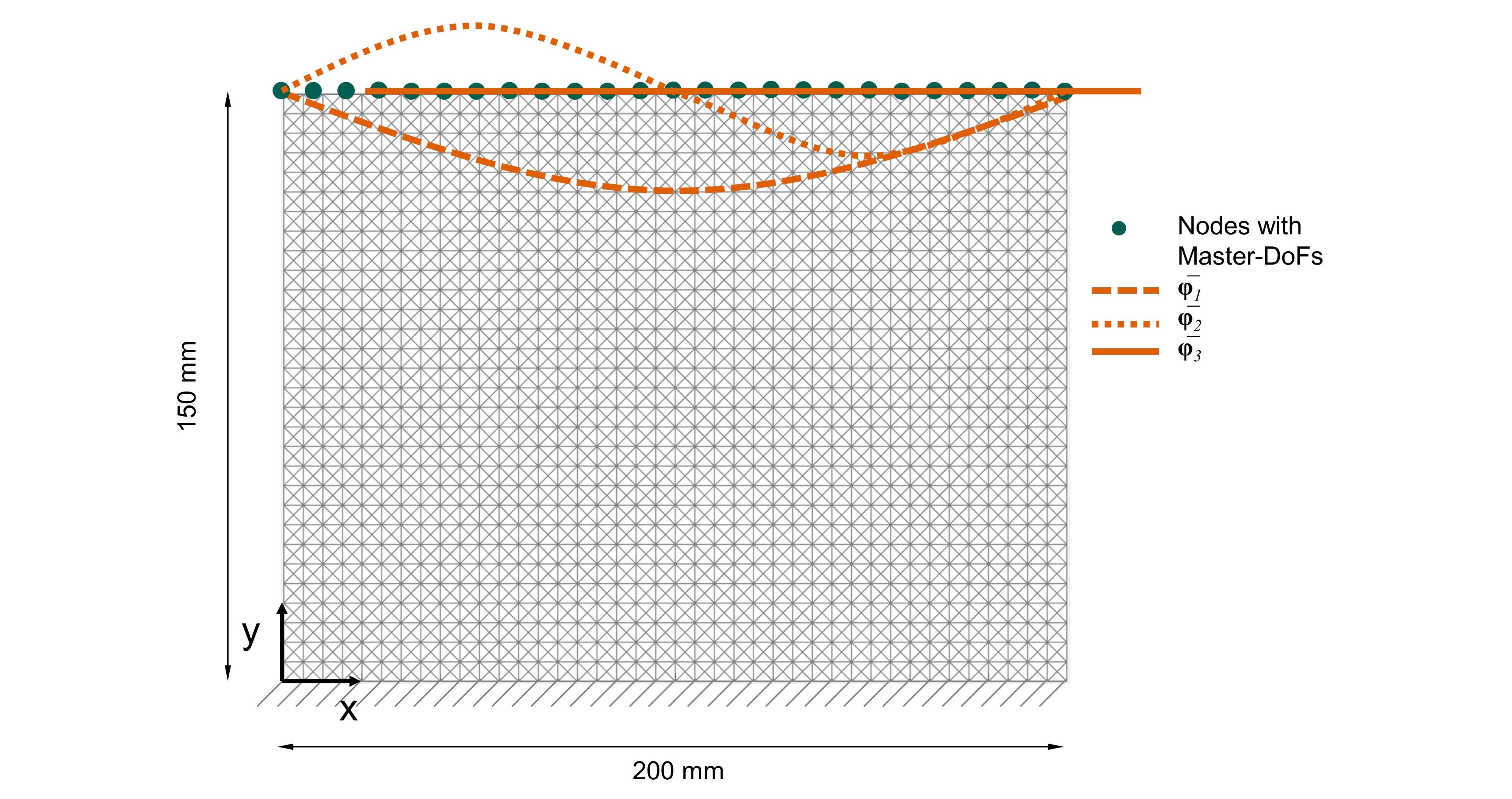}
    \caption{Shape-adaptive structure: Parameterized design space and desired deformation modes $\bm{\bar{\upvarphi}}_i$}
    \label{figure9}
\end{figure*}

\begin{figure*}[t!]
    \centering
    \includegraphics[width=16cm]{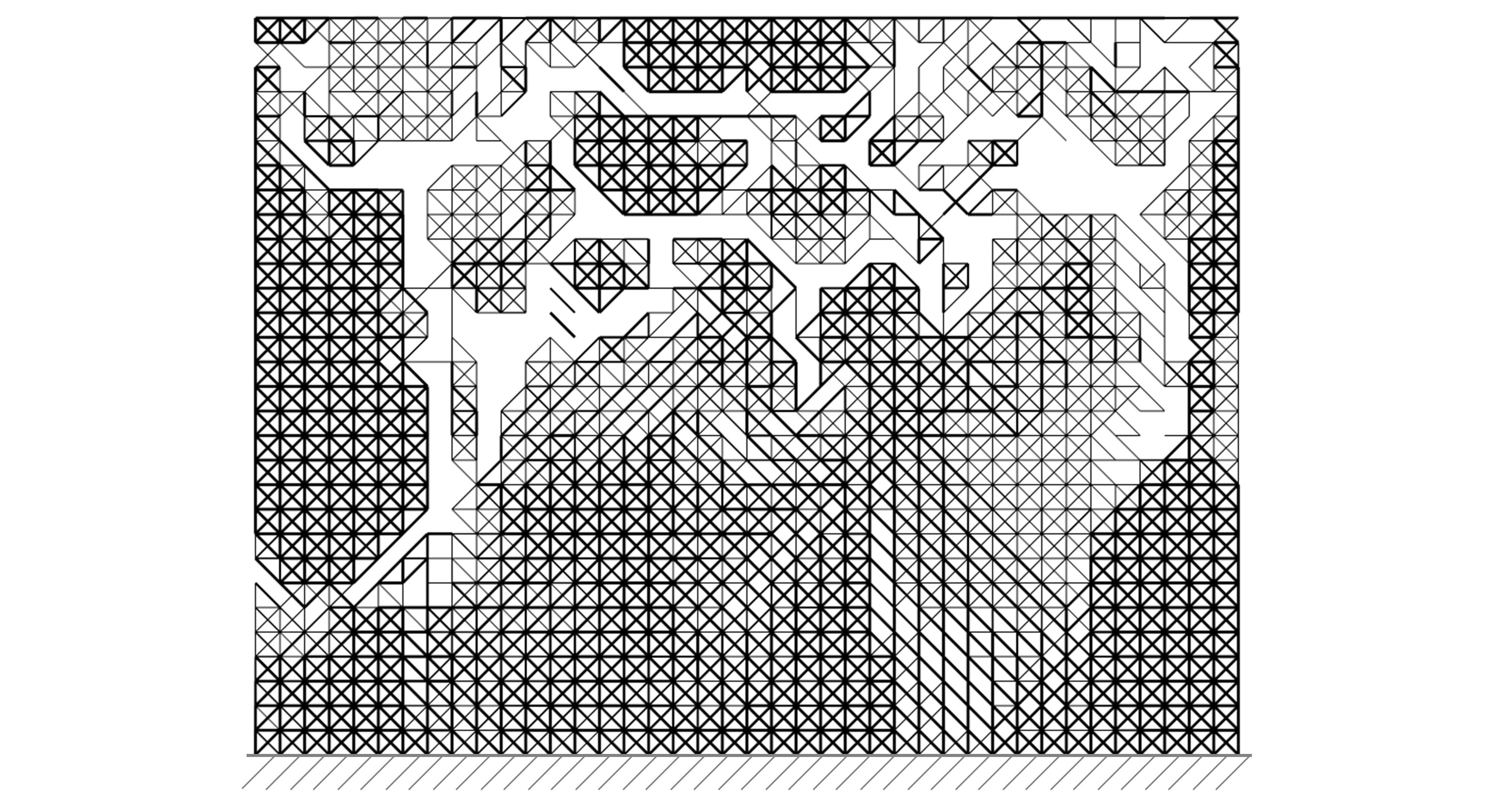}
    \caption{Shape-adaptive structure: Solution for $\mu=100$.}
    \label{figure10}
\end{figure*}

The third example is a mechanism with triple pseudo-mobility and complex kinematics. The active DoFs, the dimensions of the structure, the parameterization, as well as the desired deformation modes can be seen in \autoref{figure9}.
The parameterization consists of 4870 beam elements and the number of structural DoFs amounts to 3813, 82 of them are defined as active DoFs. These are all $x$ and $y$ displacements of the nodes on the free contour. The structure was fixed on one side. The optimization algorithm was stabilized with 20 modes. The allowed volume $V$ was 3409, which corresponds to 70 \% of the total possible volume. Five calculations each were performed with nine different values for $\mu$ , equally spaced between 100 and 900. One of the resulting designs can be seen in \autoref{figure10}. 
The eigenvalues of the first 5 eigenmodes are shown in \autoref{Table3}. As can be seen, here the first 3 eigenvalues are very similar, as desired. A selectivity of 12.1 was achieved.

\begin{table}
    \begin{center}
    \caption{Shape-adaptive structure: Eigenvalue analysis}
    \begin{tabularx}{230pt}{@{}c|ccccc|c@{}}
        \toprule
        i= & 1 & 2 & 3 & 4 & 5 & S \\
        \midrule
        $\lambda_i$ & 147.9 & 154.0 & 164.8 & 1992.2 & 3917.5 & 12.1 \\
        \bottomrule
    \end{tabularx}
    \label{Table3}
    \end{center}
\end{table}

\begin{figure*}[t!]
    \centering
    \includegraphics[width=16cm]{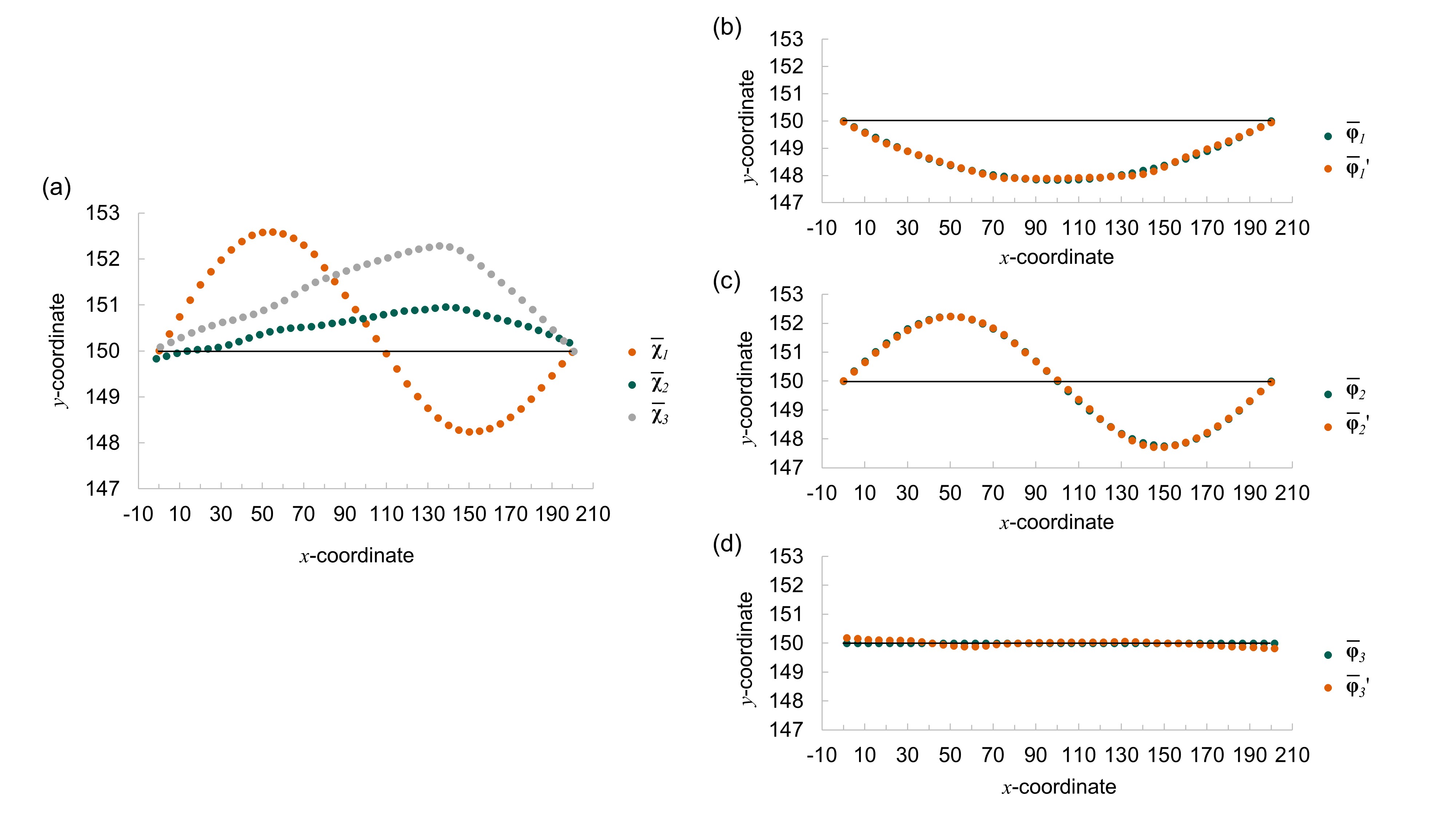}
    \caption{Shape-adaptive structure: (a) Kinematic eigenmodes $\bm{\bar{\upchi}}_i$, (b)-(d) comparison of the linear combination of kinematic eigenmodes with the desired deformation modes $\bm{\bar{\upvarphi}}'_i$}
    \label{figure11}
\end{figure*}

\begin{figure*}[t!]
    \centering
    \includegraphics[width=16cm]{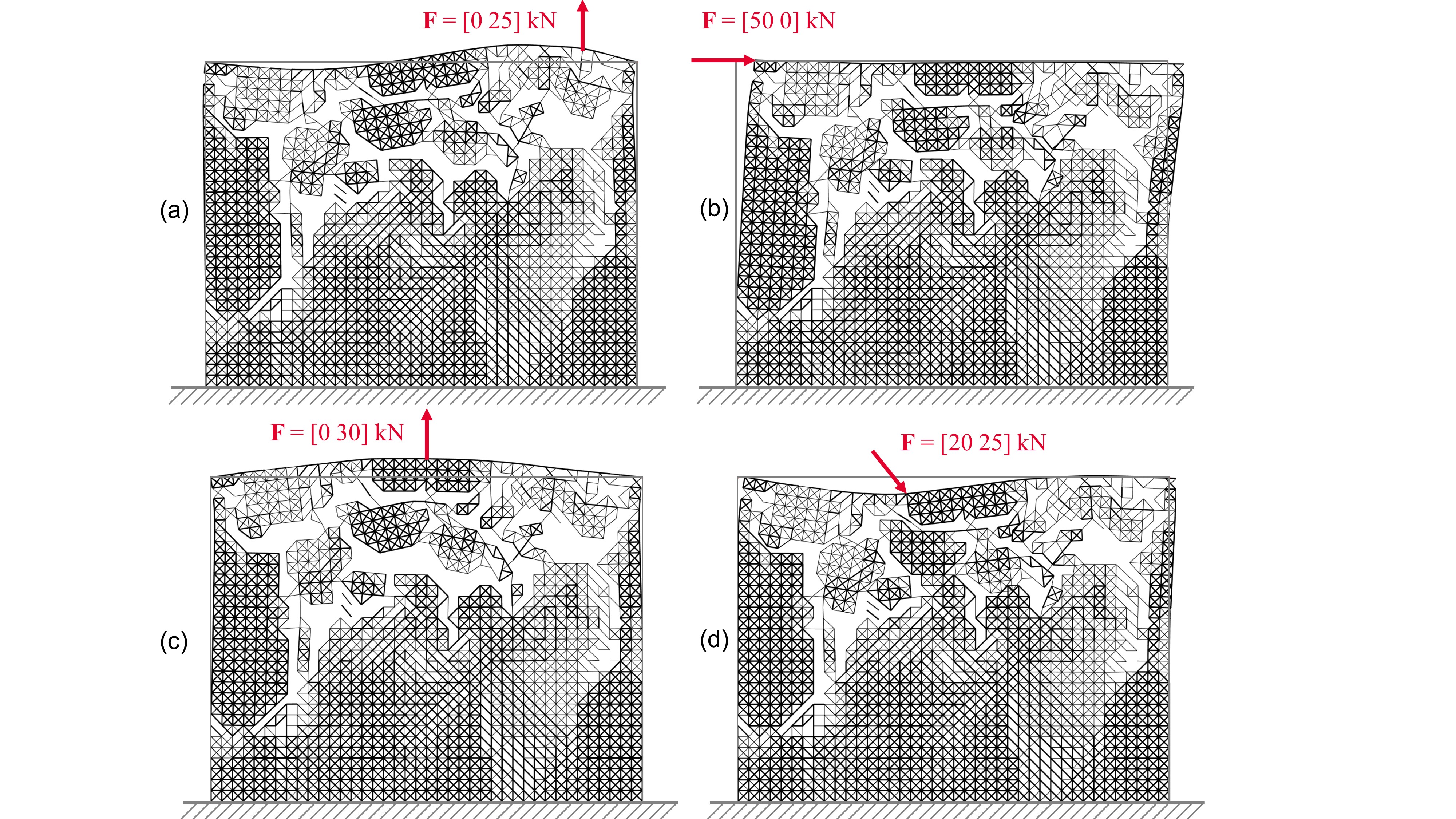}
    \caption{Shape-adaptive structure: Approximation of desired deformation modes}
    \label{figure12}
\end{figure*}

The first three eigenmodes of the structure are shown in \autoref{figure11} (a). The second and third eigenmodes show little similarity to the desired deformation modes.
However, also in this case the eigenmodes can be well approximated using a linear combination, as shown in \autoref{figure11} (b). The approximations can be achieved by the following linear combinations:
\begin{align}
    \bm{\bar{\upvarphi}}'_1= -0.26 \cdot \bm{\bar{\upchi}}_{1} - 0.35 \cdot \bm{\bar{\upchi}}_{2}  - 0.90 \cdot \bm{\bar{\upchi}}_{3} 
\end{align}
\begin{align}
    \bm{\bar{\upvarphi}}'_2= 0.96 \cdot \bm{\bar{\upchi}}_{1} - 0.13 \cdot \bm{\bar{\upchi}}_{2}  - 0.23 \cdot \bm{\bar{\upchi}}_{3} 
\end{align}
\begin{align}
    \bm{\bar{\upvarphi}}'_3= -0.04 \cdot \bm{\bar{\upchi}}_{1} - 0.92 \cdot \bm{\bar{\upchi}}_{2} - 0.38 \cdot \bm{\bar{\upchi}}_{3} 
\end{align}
The approximation here is not as good as in the previous examples. Also, the extended cosine similarity of 0.9984 is slightly lower than in the previous examples. The selectivity is also lower. In \autoref{figure12} selected load cases can be seen. In \autoref{figure12} (a) it can be seen that the second desired deformation mode (sinusoidal) can be addressed well, but a second force is needed to suppress the components of the first desired deformation mode (parabolic). In the deformation shown in \autoref{figure12} (b) it can be seen that the third desired deformation mode (horizontal displacement) can be well addressed with one force. \autoref{figure12} (c) shows the deformation according to the second desired deformation mode. \autoref{figure12} (d) shows a combination of all three desired deformation modes. Like the optimization algorithm presented in \cite{Kirmse2021}, the optimization algorithm for mechanisms with multiple pseudo-mobility is thus also suitable for the synthesis of compliant mechanisms with a large number of active DoFs.

%% file: sec_6_0_Conclusion_and_future_work.tex
\section{Conclusion and future work}
\label{sec:Conclusion and future work}

The modal approach for the synthesis of compliant mechanisms provides a load-case independent design strategy with high potential of filling the numerous gaps in the state of the art, since it inherently considers transverse load and is able to operate with a large number of output DoFs. In the paper, the extension of the modal approach to design tasks with multiple pseudo-mobility (usually referred to as design multi-DoF compliant mechanisms) is described and applied to several examples. The results show that the method is able to successfully synthetize mechanisms for a pseudo-mobility of two and three and implement a kinematics defined over a large number of DoF.
The synthesis procedure presented in this paper was developed assuming small deformations. In our future work, we intend to apply this synthesis method to the case of large deformations (geometric nonlinearity). This will require an appropriate extension of the procedure. It would be very interesting to combine this synthesis approach with motion control procedures, as already mentioned in \cite{Sachse2021}.

A MATLAB-code of the algorithm can be found at www.tu-chemnitz.de/mb/mp/multiplepseudomobility.

\section*{Funding}

This work was funded by the Deutsche Forschungsgemeinschaft (DFG, German Research Foundation ) – project number HA 7893/3–1 .